%% file: main.tex
\documentclass[10pt,shortpaper,twoside,web,logowidth=10.5]{ieeecolor}  % Comment this line out if you need a4paper

\usepackage{generic}
\usepackage{graphics} % for pdf, bitmapped graphics files
\usepackage{epsfig} % for postscript graphics files
\usepackage{amsmath} % assumes amsmath package installed
\usepackage{amssymb}  % assumes amsmath package installed
\usepackage{tensor}
\usepackage{siunitx}

\usepackage{cite}
\usepackage{url}
\usepackage{siunitx}
\usepackage{color}
\usepackage{times}

\usepackage{epsfig}
\usepackage{svg}
\usepackage{balance}
\usepackage{graphicx}
\usepackage{amsfonts}
\usepackage{fancyhdr}
\usepackage{comment}
\usepackage{changepage}
\usepackage{bm}
\usepackage{calligra}
\usepackage{tabularx}
\usepackage{mathtools}
\usepackage{arydshln}
\usepackage{latexsym} 
\usepackage{float}
\usepackage{tabstackengine}
\usepackage{siunitx}
\usepackage{colortbl}
\usepackage{hhline}
\usepackage{bookmark}
\usepackage{dirtree}
\usepackage{subfigure}
\usepackage{threeparttable}
\usepackage{siunitx}

\usepackage{xcolor}
\usepackage{booktabs} % For professional looking tables
\usepackage{siunitx} % To align numbers in the table
\usepackage{graphicx}
\usepackage{xcolor}
\usepackage{multirow}
\usepackage{lineno} %line number
\usepackage{hyperref}
\hypersetup{
    colorlinks=true,
    linkcolor=subsectioncolor,
    filecolor=magenta,      
    urlcolor=cyan,
}

\DeclareMathOperator*{\argmax}{arg\,max}
\DeclareMathOperator*{\argmin}{arg\,min}

\input{shortcuts} % shortcuts

\def\BibTeX{{\rm B\kern-.05em{\sc i\kern-.025em b}\kern-.08em
    T\kern-.1667em\lower.7ex\hbox{E}\kern-.125emX}}
\begin{document}

% \begin{linenumbers}

\title{\LARGE \bf
    PALoc: Advancing SLAM Benchmarking with Prior-Assisted 6-DoF Trajectory Generation and Uncertainty Estimation}

\author{
Xiangcheng Hu,
Linwei Zheng,
Jin Wu,
Ruoyu Geng,
Yang Yu,
Hexiang Wei,\\
Xiaoyu Tang,
Lujia Wang,
Jianhao Jiao,
and Ming Liu,~\emph{Senior Member, IEEE}
% and Ming Liu$^{*\dagger}$, \emph{Senior Member, IEEE}
% <-this % stops a space
% % \thanks{*Any organization did not support this work}% <-this % stops a space
% \thanks{$^{1}$ Robotics Institute,  
%   Department of Electronic and Computer Engineering,
%   The Hong Kong University of Science and Technology \texttt{\{xc.hu, jjiao\}@connect.ust.hk}, \texttt{\{eelium\}@ust.hk}}%
\thanks{This work was supported by Guangdong Basic and Applied Basic Research Foundation (No. 2021B1515120032), Guangzhou-HKUST(GZ) Joint Funding Program (No. 2024A03J0618), and Project of Hetao Shenzhen-Hong Kong Science and Technology Innovation Cooperation Zone(HZQB-KCZYB-2020083). (\emph{Corresponding Authors: Jianhao Jiao and Ming Liu})}
\thanks{X. Hu, J. Wu, H. Wei are with the Department of Electronic and Computer Engineering, Hong Kong University of Science and Technology, Hong Kong, China (E-mail: xhubd@connect.ust.hk)}
\thanks{L. Zheng, R. Geng, Y. Yu, L. Wang are with System Hub, Hong Kong University of Science and Technology, Guangzhou Campus, China (E-mail: eewanglj@hkust-gz.edu.cn)}
\thanks{X. Tang is with the School of Physics and Telecommunication Engineering, South China Normal University, Guangzhou, China. (E-mail: tangxy@scnu.edu.cn)}
\thanks{J. Jiao is with the Department of Computer Science, University College London, Gower Street, WC1E 6BT, London, UK. (E-mail: ucacjji@ucl.ac.uk)}
\thanks{M. Liu is with Hong Kong University of Science and Technology Guangzhou Campus, China. (E-mail: eelium@hkust-gz.edu.cn)}
}

\maketitle
% 四点贡献
% 1.system
% 2.map-degenration factor
% 3.gravity factor
% 4.轨迹量化评估
% \begin{abstract}
% 预设的
% introsuction 我们的方法，conclusion之前的discussion，essential的原因
% 引入了更多的信息量
% Unavailability of dedicated GT (GT) tracking sensors making it intractable to 

% Accurately generating GT trajectories is essential for evaluating SLAM algorithms, particularly under varying environmental conditions.

% 实验表明，在多样的campus环境中map精度至少提高30%，MOCAP室内直接轨迹ATE精度相比较ICP/NDT算法提高30%，并且鲁棒性大大提升。

\begin{abstract}
    Accurately generating ground truth (GT) trajectories is essential for Simultaneous Localization and Mapping (SLAM) evaluation, particularly under varying environmental conditions. This study presents PALoc, a systematic approach that leverages a prior map-assisted framework for the first-time generation of dense six-degree-of-freedom (6-DoF) GT poses, significantly enhancing the fidelity of SLAM benchmarks across both indoor and outdoor environments. Our method excels in handling degenerate and stationary conditions frequently encountered in SLAM datasets, thereby increasing robustness and precision.
    A critical feature of PALoc is the detailed derivation of covariance within the factor graph, enabling an in-depth analysis of pose uncertainty propagation. This analysis plays a pivotal role in illustrating specific pose uncertainty and in elevating trajectory reliability from both theoretical and practical perspectives.
    Additionally, we provide an open-source toolbox\footnote{\url{https://github.com/JokerJohn/Cloud_Map_Evaluation}} for the criteria of map evaluation, facilitating the indirect assessment of overall trajectory precision.
    Experimental results show at least a 30\% improvement in map accuracy and a 20\% increase in direct trajectory accuracy compared to the Iterative Closest Point (ICP) \cite{sharp2002icp} algorithm across diverse environments, with substantially enhanced robustness.
    Our publicly available solution, PALoc\footnote{\url{https://github.com/JokerJohn/PALoc}}, extensively applied in the FusionPortable\cite{Jiao2022Mar} dataset, is geared towards SLAM benchmark augmentation and represents a significant advancement in SLAM evaluation.
\end{abstract}

% jjiao: 几点建议
% 1. 标红的字需要重点改写一下. ``few datasets focus on how to produce high-precision ...'' 个人觉得有点绝对, 其实也有数据集讲怎么生成gt, 如oxford数据. (这个换了一种说法了)

% 我们的文章通过解答以下这几个问题, 使得我们的文章different且具有novelty:　我们生成的gt是否能生成可行度/不确定度/error的估计, 以显示这个gt的可信程度; （这个本篇文章做不完）
% 我们生成gt的方法是不是在室内外结构化/非结构化的道路上都能提供gt (这体现了我们方法的通用性)? 我们的方法是否在机器人剧烈运动的情况下也能生成可靠的gt (这体现了我们方法的鲁棒性)?  （这个主要是算法的鲁棒性，可以单独搞一篇多传感器融合的文章）
% 通过我们方法生成的gt跟专业gt的方法差多少 (这体现了我们方法的准确度)? 能否给一个详细表格比较? (这个比较目前很多数据集都提供不了, 而且很有代表意义) （这个可以）

% 2. 篇文章: https://openaccess.thecvf.com/content/CVPR2022/papers/Li_LiDARCap_Long-Range_Marker-Less_3D_Human_Motion_Capture_With_LiDAR_Point_CVPR_2022_paper.pdf, 可以参考人家一些写法和motivation. 本文也是遇到在非motion capture房间中的gt生成问题,跟他们的问题比较相似.
% 3. 方法要做到开源,不然缺少信服度
% 4. 本文可以定位成一个开源的gt生成系统或toolbox. 在写法\实验\contribution\motivation的描述上, 可以参考一些定位于系统或者toolbox的文章, 如kimera(语义建图系统),  pypose, openvins(https://udel.edu/~ghuang/iros19-vins-workshop/papers/06.pdf). 还可以提供debuging的工具描述

\section{Introduction}\label{introduction}
%  motivation, 3段， interesting, hard, solvable
% interesting原因：
%  - 数据集是在同一场景中评估和比较slam算法的性能的重要数据，不https://github.com/lee-zq/3DUNet-Pytorch少数据的评估gt不够准确或者没有gt，可能误导算法评估工作
%  - 室内场景只能在mcp系统下有轨迹真值，毫米级。室外系统GNSS并不稳定，无法保证所有的gt pose准确。很多数据集在这些情况下要么没有提供轨迹，要么提供轨迹不完全可靠。没有gt，对于定位算法评估而言毫无意义。
% 
% hard
%   - 在非tracking系统下，很难直接测量得到传感器的运动姿态，只能通过估计的方法得到一个相对不错的轨迹精度。
%   - 数据集的真值采集工具不同，很难统一起来，有的是莱卡采集的gt map，可以评估地图，无法评估定位，有的是全站仪采集的绝对位置，无法评估姿态。但很多评估工具提供的都是位姿轨迹评估，制作轨迹真值是必要的。
%   - 纯定位方法
%
% solvable 
% - 当前很多SOTA的slam算法基于多传感器融合的方案能够得到很高的精度，室内位置的平均距离误差甚至保持在3cm左右，准确性非常高，接近激光雷达的测量噪声水平。
%  - slam系统存在drift，但在局部drift很小，位姿估计比较准确， 尽管有闭环检测算法消除，但没有绝对先验信息的引入还是不可信。
%  - 绝对的先验信息比如不管是来源于先验点云地图还是GNSS，虽然被很好地全局一致性，但不能保证局部轨迹的平滑和准确性，有一定的噪声或者误差，这显然是不符合运动学规律的。
% - 将绝对定位信息结合局部的slam轨迹，能够很好的取长补短。这种方法在实时定位系统中出现过一些，但很少有人把它运用于制作数据集真值轨迹。
% 
%%%%%%%%%%%%%%%%%%%%%%%%%%%%%%%%%%%%%%%%%%%%%%%%%%%%%%%%%%%%%%%%%%%%%%%%%%%%%%%%%%%%%%%%

\begin{figure}[t]
    \setlength{\subfigcapskip}{-0.1cm}
    \setlength{\subfigbottomskip}{-0.1cm}
    \subfigure[ ]{
        \label{fig:corridor_normal}
        \includegraphics[width=.42\linewidth]{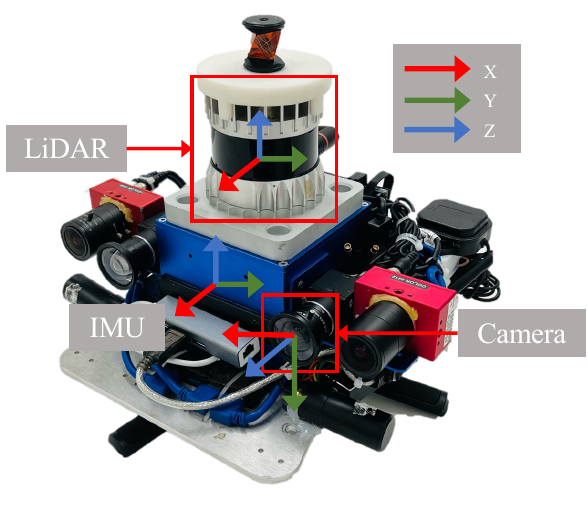}
    }
    \subfigure[ ]{
        \label{fig:corridor_degenration}
        \includegraphics[width=.48\linewidth]{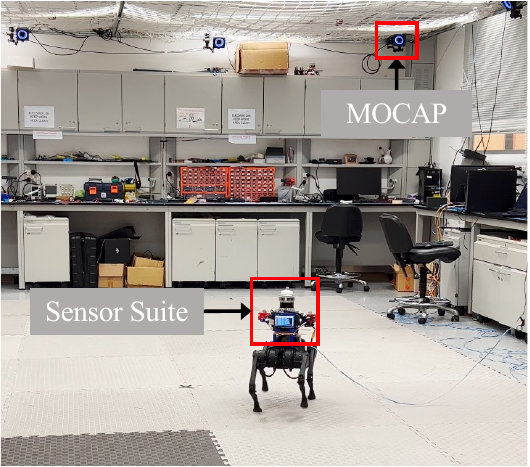}
    }
    \subfigure[ ]{
        \label{fig:corridor_degenration}
        \includegraphics[width=.95\linewidth]{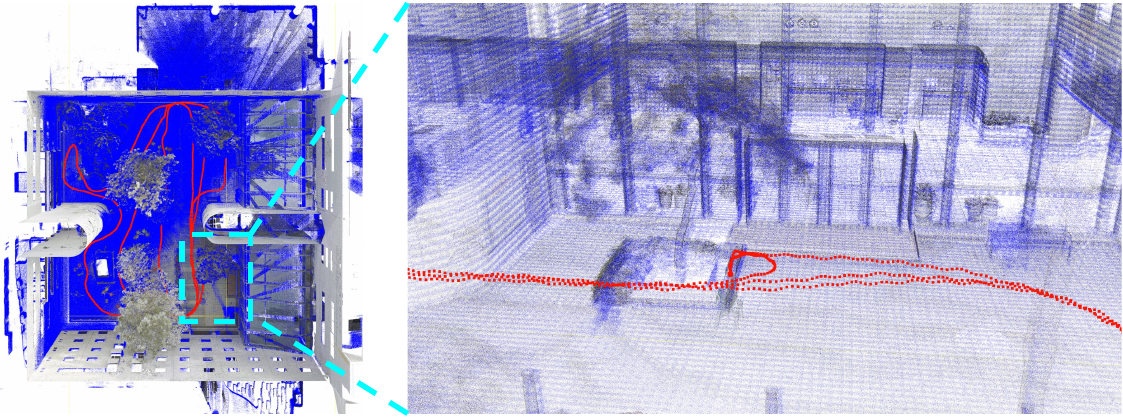}
    }
    \caption{\begin{small}(a) Sensor configuration with corresponding coordinate frames for the SLAM benchmarking. (b) Quadruped robot equipped with sensor suite in Motion Capture Room (MCR). (c) Prior RGB point cloud map with the estimated trajectory (red line) and map (blue point cloud) by PALoc on \texttt{garden\_day}.\end{small}}
    \label{fig:sensor_kit}
    \vspace{-2em}
\end{figure}

\subsection{Motivation and Challenges}
% \IEEEPARstart{G}T trajectory generation, a cornerstone in evaluating the performance of SLAM algorithms, embodies a blend of intricate geometric processing and algorithmic precision. Its essence lies in creating a benchmark against which the fidelity of SLAM algorithms can be rigorously tested. The value of GT trajectories extends beyond mere coordinates; it's about constructing a reliable canvas\cite{Geiger2012} where the nuances of navigation and mapping algorithms can be critically examined and refined. Its significance is particularly pronounced in robotics and autonomous systems, where even minor inaccuracies in SLAM evaluations can lead to substantial deviations in real-world applications\cite{Jeong2019Apr}. This underlines the profound impact and interest in developing advanced GT generation methods.

\IEEEPARstart{G}{T} trajectory generation is crucial for assessing SLAM. It requires sophisticated geometric computations to fully explore the capabilities of GT estimation methods.
Its primary goal is to establish a benchmark for thoroughly assessing the fidelity of SLAM technologies. GT trajectories offer more than just positional data; they provide a robust framework for the critical analysis and enhancement of navigation and mapping techniques. 
This is particularly vital in the fields of robotics and autonomous systems, where slight errors in SLAM assessments can result in significant real-world discrepancies. The development of GT trajectory generation methods is therefore of critical importance, reflecting its significant impact and the keen interest it generates within the community \cite{Geiger2012,Jeong2019Apr}.

Despite their centrality in SLAM evaluation, existing GT trajectory generation methods face notable limitations. Tracking-based approaches, relying on expensive instruments like motion capture system (MOCAP), Total Station (TS), Global Navigation Satellite System (GNSS), and Inertial Navigation Systems (INS), are hindered by limited environmental coverage\cite{Burri2016Jan} and occlusion susceptibility\cite{Delmerico2019May}, often resulting in trajectories with unbounded error\cite{Geiger2012, Jeong2019Apr}, limited degrees of freedom (e.g., 3-DoF)\cite{nguyen2022ntu}, and sparse tracking (e.g., 1Hz)\cite{Zhang2022Aug}. 
Scanning-based methods, while innovative in using laser scanners for prior map construction and subsequent dense 6-DoF trajectory generation through point cloud localization, grapple with poor local accuracy\cite{ramezani2020newer}, localization failures\cite{Sier2022Oct} during intense movements, and inability to handle degenerate scenarios\cite{Zhang2022Aug}. Most critically, these methods lack theoretical analysis and quantitative results, leaving a gap in measuring overall trajectory quality in practical applications.

% Despite their importance in SLAM evaluations, current GT trajectory generation methods encounter significant challenges. 
% Methods based on tracking, which utilize expensive instruments such as Motion Capture Systems (MOCAP), Total Station (TS), Global Navigation Satellite Systems (GNSS), and Inertial Navigation Systems (INS), face obstacles including restricted environmental range \cite{Burri2016Jan} and vulnerability to occlusions \cite{Delmerico2019May}. 
% These limitations can lead to trajectories characterized by potentially unlimited error margins \cite{Geiger2012, Jeong2019Apr}, constrained degrees of freedom (for instance, 3-DoF) \cite{nguyen2022ntu}, and infrequent tracking instances (such as 1Hz) \cite{Zhang2022Aug}. 
% On the other hand, scanning-based methods leverage laser scanners for initial map creation and dense 6-DoF trajectory establishment via point cloud localization but struggle with low local precision \cite{ramezani2020newer}, failures in localization amid rapid movements\cite{Sier2022Oct}, and challenges in addressing degenerate situations \cite{Zhang2022Aug}. 
% A critical shortfall of these techniques is their lack of comprehensive theoretical analysis and quantitative evaluation, creating a void in accurately determining trajectory accuracy and quality in real-world settings.

% 对应limitation 最相近的方法上的描述，概述方法的好处
% 从大的角度上说，从生物等领域引进过来，overview方法上的好处
To address these challenges, our proposed factor graph-based methodology considers temporal sensor information to ensure robustness, local accuracy, and consistency across various platforms. We have integrated specialized modules for degeneracy handling and Zero Velocity Update (ZUPT), refining trajectory accuracy and ensuring robustness against environmental dynamics. Our estimation of pose uncertainty for each frame offers theoretical insights and practical implications for trajectory generation tasks. Additionally, by integrating a quantitative evaluation module inspired by Multi-View Stereo (MVS) or other reconstruction techniques \cite{xu2020planar, Aanaes2016Nov, seb2018chall}, we offer a nuanced assessment of the generated trajectories. This comprehensive approach not only resolves existing challenges in GT trajectory generation methods but also establishes a new benchmark in this domain, demonstrating the practical feasibility and theoretical robustness of our method.
%%%%%%%%%%%%%%%%%%%%%%%%%%%%%%%%%%%%%%%%%%%%%%%%%%%%%%%%%%%%%%%%%%%%%%%%%%%%%%%%%%%

\subsection{Contribution}
In this paper, we present a comprehensive method for generating high-quality GT trajectories to augment SLAM benchmark datasets. Our key contributions are as follows:
\begin{itemize}
    \item We introduce a prior-assisted localization system to generate dense, 6-DoF trajectories for SLAM benchmarking, incorporating degeneracy-aware map factors and ZUPT factors to boost robustness and precision.
    \item We conduct a detailed analysis of uncertainty propagation within our factor graph system, estimating the uncertainty of each pose, which informs the quality and robustness of the final trajectories, offering theoretical and practical insights for trajectory generation tasks.
    \item We offer an open-source toolbox for map evaluation criteria, acting as indirect precision indicators of the generated trajectories, emphasizing the practical utility and adaptability of our method in diverse applications.
\end{itemize}

Leveraging a loosely coupled fusion approach based on the factor graph, our method swiftly adapts to state-of-the-art (SOTA) LiDAR-centric SLAM systems like LiDAR Odmetry, LiDAR-Inertial Odometry (LIO), and LiDAR-Visual-Inertial Odometry (LVIO), demonstrating versatility, scalability, and compatibility. Applied extensively in the FusionPortable dataset\cite{Jiao2022Mar}, our method successfully produced trajectories for $\bm{13}$ of $16$ sequences, achieving at least a $\bm{30\%}$ improvement in map accuracy and a $\bm{20\%}$ increase in direct trajectory accuracy compared to the ICP\cite{sharp2002icp} algorithm across diverse campus environments. To our knowledge, this approach represents the first open-source solution designed specifically for crafting 6-DoF GT trajectories in benchmarking, marking a significant contribution to SLAM research.

% \begin{itemize}
%     \item We introduce a prior-assisted localization system utilizing a factor graph to generate dense, 6-DoF GT trajectories for SLAM benchmarking, integrating degeneracy-aware map factors and Zero Velocity Update (ZUPT) factors to enhance robustness and precision.

%     % 分析对于最终精度和鲁棒性很重要
%    \item \rt{Our approach includes deriving the covariance for each factor, enabling thorough analysis of pose uncertainty propagation, thereby providing theoretical and practical insights for trajectory generation tasks.}

%     % evaluation的完整的toolbox来评估整体的地图质量，间接反应轨迹质量
%     \item We provide an open-source toolbox for map evaluation criteria that serve as indirect indicators of trajectory precision, emphasizing the practical utility and adaptability of our method in diverse applications.
% \end{itemize}
% % 提一下可扩展性等，给一个具体提升精度的指标，开源，在fp里面已经应用,支持多少sequence，兼容多少算法等定量数字
% % 松耦合,兼容性,可拓展性vertsial,extensiable
% \rt{To our knowledge, this approach is the first open-source solution designed for creating 6-DoF GT trajectories in benchmarking, signifying a notable advancement in SLAM research}.

\section{Related Work}\label{sec:related_work}
Generating  GT trajectory is a pivotal task in constructing SLAM benchmark. This section briefly reviews  GT trajectory generation methods within SLAM datasets.

A category of tracking-based GT trajectory generation methods heavily relies on tracking sensors,  offering high accuracy but fraught with several drawbacks. Datasets like\cite{Geiger2012, Huang2010Nov, Peynot2010Nov, Pandey2011Mar, Maddern2016Nov} primarily employ GNSS/INS, facing challenges like limited satellite visibility and the multi-path effect, which significantly reduce accuracy in outdoor urban areas and are not applicable for indoor dataset creation. In contrast, \cite{Burri2016Jan, Schubert2018Oct, Jiao2022Mar} employ MOCAP to generate 6-DoF GT trajectories. However, their dependency on MOCAP limits data collection to its coverage area, limiting the diversity of capturable scenarios. TS\cite{nguyen2022ntu}, while precise, encounter low pose frequency and are limited to positional tracking, struggling with coverage and occlusions.

Scanning-based methods utilize sensor measurements and prior maps \cite{ye2020monocular}, introducing LiDAR measurements noise and map noise but enabling GT trajectory creation in any area of a mapped scene. Innovations like\cite{ramezani2020newer} employ Laser Scanners for map construction, typically using brute-force NDT\cite{magnusson2007scan} or ICP\cite{sharp2002icp} algorithms for point cloud localization. While these methods have good global accuracy, they often yield sparse GT poses and lack robustness and local accuracy, particularly in LiDAR-degenerate environments. Distributed algorithms \cite{doostmohammadian2013genericity, doostmohammadian2021distributed} also contribute significantly to estimation, and target localization. LIO/LVIO methods \cite{Zhang2014Jun, Zhang2015May, xu2022fast, ye2019tightly, Qin2020May, jiao2021robust, huang2021bundle,lin2022r} estimate poses using temporal sensor data, handling scene changes and intense movements with high local accuracy but limited global precision, and are rarely used for GT trajectory generation. In large outdoor scenarios, the fusion of LIO/LVIO with GNSS/INS has been employed for GT generation\cite{Jiao2022Mar}, overcoming conventional GNSS/INS limitations.

% traditional词有异议，
% some xx propose，局部精度高，全局精度比较局限，并没有应用到GT生成任务中
% 分析ndt等方法无法保证局部精度，而LIO能保证但全局精度比较局限，因此our method
% 方法的细节上的区别，怎么改进的
Following the scanning-based method, our approach further integrates LIO/LVIO within a unified factor graph-based fusion framework. This enables our method to achieve both excellent local precision and global consistency, even in dynamic movements. To enhance robustness and accuracy, we implement degeneracy-aware map factors and ZUPT factors. We offer map evaluation benchmarks for succinct trajectory precision indication, combined with theoretical analysis of pose uncertainty propagation within our system. By estimating the uncertainty of each pose, we provide clear insights into pose quality, thus guiding trajectory generation tasks from both theoretical and practical perspectives. This comprehensive method differentiates our approach in the SLAM dataset landscape, marking a significant departure from existing practices.

%%%%%%%%%%%%%%%%%%%%%%%%%%%%%%%%%%%%%%%%%%%%%%%%%%%%%%%%%%%%%%%%%%%%%%%%%%%%%%%%%%%%%%

% 给实物或者实验结果图
\begin{figure*}
    \centering
    \includegraphics[width=0.85\textwidth]{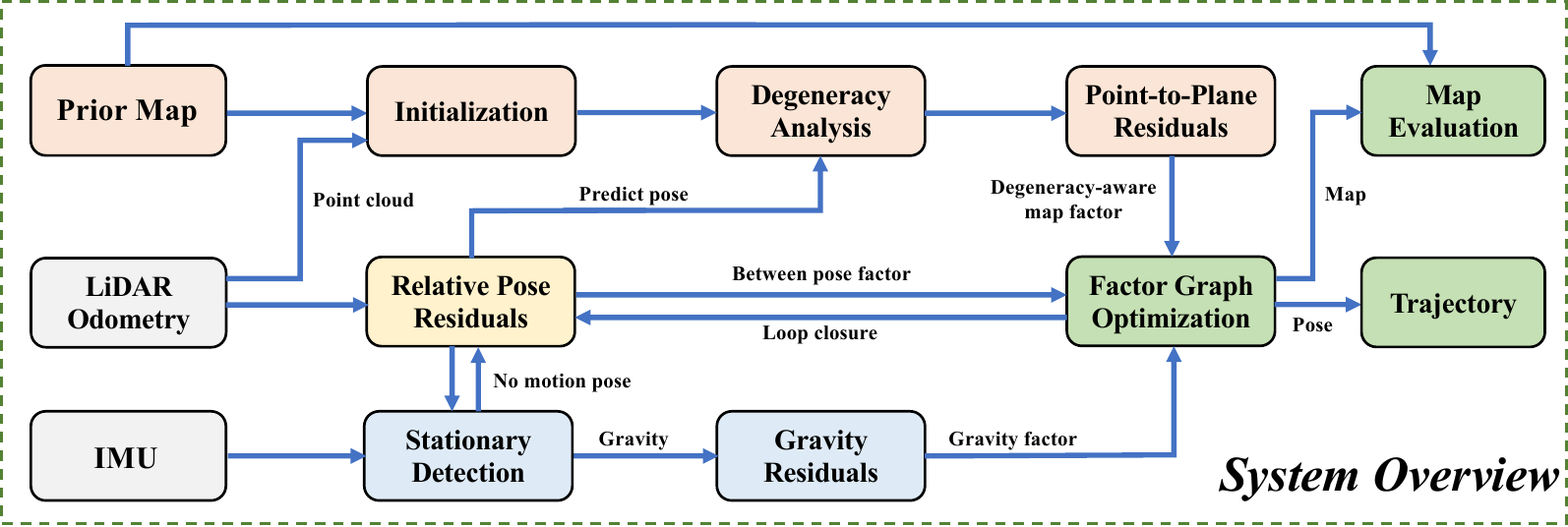}
    \caption{\begin{small} \textbf{System Pipeline Overview}. This figure illustrates the architecture of our system, starting with initialization on a prior map and LiDAR odometry. Degeneracy analysis and point-to-plane registration are employed to create a degeneracy-aware map factor. The system also processes odometry and IMU data for stationary detection, forming no motion factors, and integrating gravity factors. Following the optimization of the factor graph on a frame-wise basis, loop closure detection is carried out, contributing to the Loop factor. This sequential pipeline results in the generation of estimated poses and maps, which assists in the indirect evaluation of trajectory accuracy. \end{small}}
    \label{fig:pipeline}
    \vspace{-1.5em}
\end{figure*}

\section{Problem statement and Formulation} \label{sec:problem_statement}
In this section, we formulate the trajectory generation task as a localization problem. Our method employs a factor graph framework, accounting for various factors essential for accurate trajectory estimation in complex environments.

\subsection{Notations and Definitions}\label{sec:notations}

For clarity and precision in our problem formulation, we adopt a consistent notation system. The Inertial Measurement Unit (IMU) is rigidly attached to the LiDAR sensor, defining the body frame \( \bm{B} \) which serves as the reference frame for the system. The world frame is denoted by \( \bm{W} \), and the LiDAR frame is denoted by \( \bm{L} \). The robot pose at any discrete time instance \( k \) is denoted by \( \bm{X}_k \), comprising its position, \( \bm{t}_k \in \mathbb{R}^3 \), and orientation, \( \bm{R}_k \in \mathrm{SO}(3) \). Velocity, accelerator biases, and gyroscope biases are denoted by \( \bm{v}_k \), \( \bm{b}_{a,k} \), and \( \bm{b}_{\omega,k} \), respectively, but are not included in the state variable as our focus is on pose graph. The state vector \( \bm{\mathcal{X}} \) is thus defined in terms of the pose: $\bm{\mathcal{X}} = \{ \bm{X}_1, \ldots, \bm{X}_P \}$, where \( P \) represents the total number of time steps. The environmental context is provided by the prior map \( \mathcal{M} \), with \( \mathcal{M}_k \) referencing the relevant subset at the initial pose \( \bm{x}_k \). Observations at time \( k \) incorporate LiDAR \( \mathcal{L}_k \) and IMU measurements \( \mathcal{I}_k = \{ \bm{a}_k, \bm{\omega}_k \} \), collectively represented as: $\mathcal{Z}_k = \{ \mathcal{M}_k, \mathcal{L}_k,  \mathcal{I}_k \}$.

%顶格
\begin{figure}[t]
    \centering
    \includegraphics[width=0.35\textwidth]{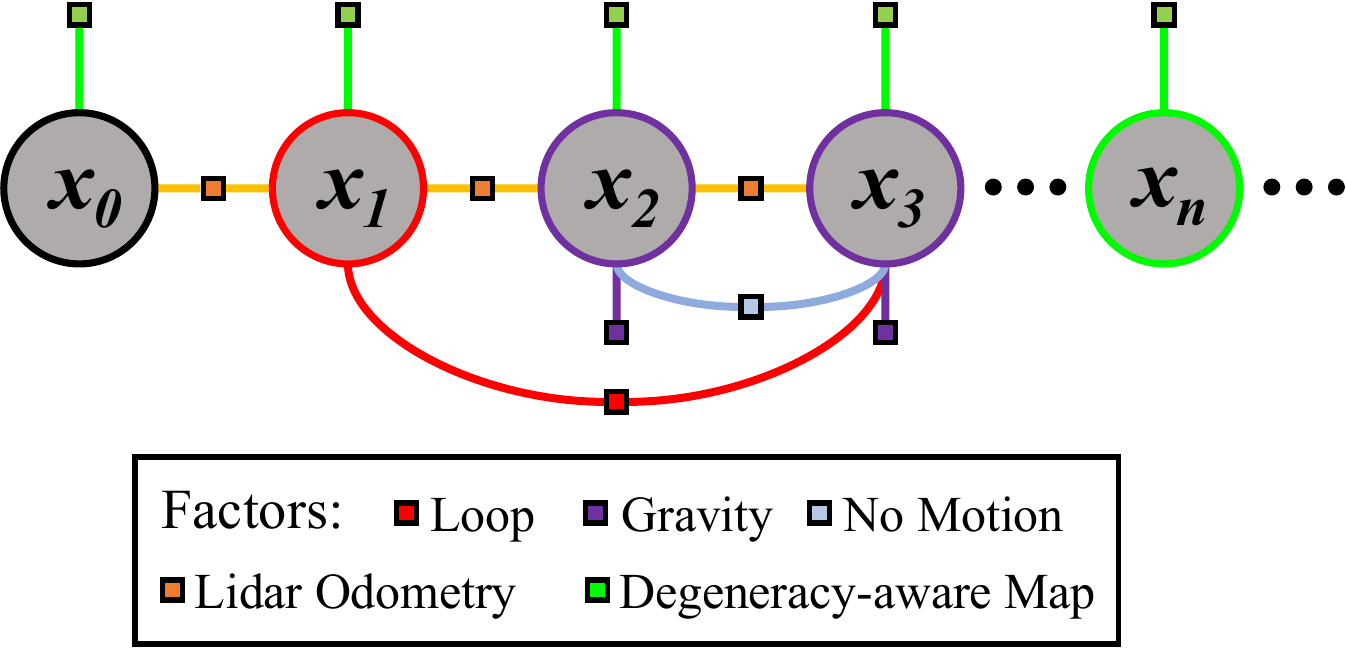}
    \caption{
        \begin{small}\textbf{Factor Graph of Our Proposed System}. Gray circles represent different system states at specific times, and colored rectangles symbolize various factors. The purple rectangle signifies the gravity factor, connected to states outlined in purple, illustrating gravity constraints during stationary periods. States with green outlines are indicative of LiDAR degenerate scenarios, and those with red outlines mark instances of successful loop closure detection. An in-depth discussion on uncertainty propagation is detailed in Section \ref{sec:uncertainty_analysis}.\end{small}}
    \label{fig:factor_graph}
    \vspace{-1.5em}
\end{figure}

\subsection{Maximum-A-Posteriori Estimation}\label{sec:map_optimization}
The cornerstone of our state estimation is the maximum a posteriori (MAP) framework, which aims to identify the most likely state variables \( \bm{\mathcal{X}} \). Given the observations \( \mathcal{Z} \), this process is described by:
\begin{small}
    \begin{equation*}
        \begin{aligned}
            \hat{\bm{\mathcal{X}}} & = \argmax_{\bm{\mathcal{X}}} \left[ p(\bm{\mathcal{X}} | \mathcal{Z}) \right] & = \argmax_{\bm{\mathcal{X}}} \left[ p(\mathcal{Z} | \bm{\mathcal{X}}) p(\bm{\mathcal{X}}) \right],
        \end{aligned}
    \end{equation*}
\end{small}
where \( p(\mathcal{Z}) \) is a constant.

Under the conditional independence assumption, the joint likelihood decomposes into a product of individual probabilities as (\ref{eq:joint_likely}).
\begin{small}
    \begin{equation}\label{eq:joint_likely}
        \hat{\bm{\mathcal{X}}} = \argmax_{\bm{\mathcal{X}}} \left[ p(\bm{\mathcal{X}}) \prod_{k=1}^{P} \prod_{i=1}^{M} p(\bm{z}_i^k | \bm{X}_k, \bm{\mathcal{X}}) \right],
    \end{equation}
\end{small}
where \( P \) and \( M \) denote the number of time steps and factors. (\ref{eq:joint_likely}) is equivalent to minimizing the sum of squared residuals across all factors:
\begin{equation*}
    \hat{\bm{\mathcal{X}}} = \argmin_{\bm{\mathcal{X}}} \left[ \bm{\Sigma}_{k=1}^{P} \bm{\Sigma}_{i=1}^{M} \bm{r}_i^k(\bm{\mathcal{X}})^\top \bm{\Sigma}_i^{-1} \bm{r}_i^k(\bm{\mathcal{X}}) \right].
\end{equation*}
The optimization of \( \bm{\mathcal{X}} \) is achieved using nonlinear least squares techniques, such as Gauss-Newton methods, accommodating the inherent uncertainties in sensor data.

\subsection{Factor Graph Formulation}\label{sec:formulation}
In our system, we represent the factor graph as $\mathcal{G} = (\mathcal{X}, \mathcal{F}, \mathcal{E})$, forming the backbone of our pose estimation framework. Here, $\mathcal{X}$ comprises the set of state variables related to robot poses, $\mathcal{F}$ embeds constraints between these variables, and $\mathcal{E}$ indicates the connections between factors and variables. A graphical illustration of our factor graph, integrating five distinct factor types, is shown in Fig.\ref{fig:factor_graph}. Detailed explanations of these factors are provided as follows:

\subsubsection{Lidar-based Odometry Factor (LO)}
This factor imposes constraints on consecutive robot poses $\bm{X}_k$ and $\bm{X}_{k-1}$, aligning them based on odometry measurements. The intricate details of this factor are elaborated in Section \ref{sec:odom_cov}.
\subsubsection{Loop Closure Factor (LC)}
The loop closure factor aims to enforce congruence between two robot poses, $\bm{X}_i$ and $\bm{X}_j$, which, despite being temporally distant, should be spatially proximal due to a loop closure event.
% 是否需要整合
\subsubsection{No Motion Factor (NM)}
Designed to cater to stationary periods, the no motion factor\cite{Rosinol20icraKimera} ensures that the relative pose remains constant during such intervals. Further insights into this factor are provided in Section \ref{sec:no_motion}.
\subsubsection{Gravity Factor (GF)}
This factor critically influences the estimation process under ZUPT conditions by constraining the magnitude of the gravity vector. A detailed formulation of this factor can be found in Section \ref{sec:ngf}.

\subsubsection{Degeneracy-Aware Map Factor (DM)}
The degeneracy-aware map factor constrains the robot pose with a prior map, thus ensuring robustness in degenerate scenarios. Detailed discussions on this factor are presented in Section \ref{sec:damf}.

\section{System Overview}\label{sec:system_overview}
We develop our system under a set of simplifying assumptions to streamline the design process:
\begin{enumerate}
    \item The LiDAR and IMU are hardware time-synchronized and well-calibrated, ensuring precise data alignment.
    \item Our research utilizes Pose SLAM\cite{Ila2010Info}, where only poses are included in the graph, ensuring efficiency and optimal map construction.
\end{enumerate}

Fig.\ref{fig:pipeline} presents an overview of our system's architecture, with the caption detailing its components and workflow.

\section{ZUPT Factors}\label{sec:ngf}
Inspired by previous gravity and ZUPT work \cite{Kubelka2022Mar, ye2019tightly, qin2019vins, Rosinol20icraKimera}, we implement the gravity factor and no motion factor to handle these frequently occurred static scenes.

\subsection{Stationary Detection}
The system needs to pass a two-stage evaluation to be seen as static by analyzing the IMU data and LIO poses. Given a set of recent accelerometer measurements at time $i$, $\left\{ \bm{a}_i, \bm{\omega}_i \right\}$, we initially compute the mean values and variations of these measurements as $\overline{\boldsymbol{a}}=\frac{1}{N} \bm{\Sigma}_{i=1}^N \boldsymbol{a}_i$ and $\overline{\boldsymbol{\omega}}=\frac{1}{N} \bm{\Sigma}_{i=1}^N \boldsymbol{\omega}_i$.
% \begin{equation*}
% \begin{gathered}
% \overline{\boldsymbol{a}}=\frac{1}{N} \Sigma_{i=1}^N \boldsymbol{a}_i, \quad \overline{\boldsymbol{\omega}}=\frac{1}{N} \Sigma_{i=1}^N \boldsymbol{\omega}_i \\
% \end{gathered}.
% \end{equation*}
Firstly, we establish thresholds $\bm{\epsilon}_a$ and $\bm{\epsilon}_\omega$ for the variations of acceleration and angular velocity.  If $\Delta\boldsymbol{a} < \bm{\epsilon}_a$ and $\Delta\boldsymbol{\omega} < \bm{\epsilon}_\omega$, the system is tentatively classified as static as (\ref{eq:acc_thres}).
\begin{equation}\label{eq:acc_thres}
    \Delta \bm{a}=\max _i\left\|\boldsymbol{a}_i-\overline{\boldsymbol{a}}\right\|, \quad \Delta \boldsymbol{\omega}=\max _i\left\|\boldsymbol{\omega}_i-\overline{\boldsymbol{\omega}}\right\|.
\end{equation}
Next, we examine the LIO pose, characterized by the relative translation $\boldsymbol{t}_{i, i-1}$ and rotation $\boldsymbol{R}_{i, i-1}$. We set thresholds $\bm{\epsilon}_t$ and $\bm{\epsilon}_R$ for the variations in translation and rotation, respectively. If both $|\boldsymbol{t}_{i, i-1}| < \bm{\epsilon}_t$ and $|\angle(\boldsymbol{R}_{i, i-1})| < \bm{\epsilon}_R$, we further confirm the static state. Our refined two-stage approach offers a consistent method for detecting ZUPT conditions.
% 画个图
\subsection{Gravity Factor }
We aim to estimate the pose \(\bm{p} \in SE(3)\), given the gravity vector \(\bm{g} \in \mathbb{R}^3\) is a normalized constant. The measured acceleration in the body frame, \(\bm{a}_m^b\), when stationary, is an approximation of the negative gravity vector. This acceleration is transformed into the world frame \(\bm{a}^w\) by the rotation matrix \(\bm{R}\) encapsulated in the pose \(\bm{p}\),
$\bm{a}^w = \bm{R} \bm{a}_m^b$. The error \(\bm{r}_{\text{gf}}\) is then defined as the deviation of the normalized world frame acceleration from the fixed gravity direction $\bm{r}_{\text{gf}} = \bm{a}^w / {\|\bm{a}^w\|} - \bm{g}$.
% \begin{equation}
% \bm{r}_{\text{gf}} = \frac{\bm{a}^w}{\|\bm{a}^w\|} - \bm{g},
% \end{equation}
% \begin{equation}
% \bm{r}_{\text{gf}} = \bm{a}^w / {\|\bm{a}^w\|} - \bm{g},
% \end{equation}
The Jacobian matrix \(\bm{J}_{\text{gf}}\) with respect to the pose is given by (\ref{eq:gf_Jaco}).
\begin{equation}\label{eq:gf_Jaco}
    \bm{J}_{\text{gf}} = \left[\begin{array}{cc}
            \bm{0}_{3 \times 3} & \frac{\bm{R}\left[\bm{a}_m^b\right]_{\times}}{\|\bm{a}^w\|}
        \end{array}\right]_{3\times6},
\end{equation}
where \(\left[\bm{a}_m^b\right]_{\times}\) denotes the skew-symmetric matrix of the vector \(\bm{a}_m^b\). The covariance matrix \( \bm{\Sigma_{}} \) is then derived from the inverse of the Fisher information matrix \(\bm{H}\) and is obtained as follows:
\begin{equation*}
    \bm{H} = \bm{J}_{\text{gf}}^\top \bm{W}_{\text{gf}} \bm{J}_{\text{gf}}, \quad \bm{\Sigma}_{\text{gf}} \approx \bm{H}^{-1}.
\end{equation*}
where \(\bm{W}_{\text{gf}}\) is the weight matrix associated with IMU noise.

\subsection{No Motion Factor}
\label{sec:no_motion}
The no motion factor can be  defined as $\bm{r}_{\text{nm}} = \text{Log}(\bm{X}_{t}^{-1} \bm{X}_{t-1})$, where \( \bm{X}_{t} \) and \( \bm{X}_{t-1} \) denote the consecutive poses, and \( \text{Log} \) translates SE(3) into its Lie algebra. The covariance matrix \( \bm{\Sigma}_{\text{nm}} \) is depicted as a diagonal matrix with small entries, indicating minimal uncertainty in both rotation and translation.

% \begin{equation}
%     \bm{\Sigma}_{\text{nm}} = \text{diag}(\sigma_{\text{rot}}^2, \sigma_{\text{rot}}^2, \sigma_{\text{rot}}^2, \sigma_{\text{trans}}^2, \sigma_{\text{trans}}^2, \sigma_{\text{trans}}^2).
% \end{equation}
% where \( \sigma_{\text{rot}}^2 \) and \( \sigma_{\text{trans}}^2 \) are the variances associated with rotation and translation uncertainties.

\section{Degeneration-aware Map Factor}
\label{sec:damf}
Inspired by \cite{Jiang2022Puma, tagliabue2021lion, Tuna2022NovXicp, jiao2021Greedy, Zhang2016On, Zhang2014Jun}, Our proposed DM approach augments the traditional point-to-plane ICP algorithm by integrating degeneracy detection and uncertainty estimation modules.

\subsection{Point-to-Plane ICP}
The point-to-plane ICP residual is expressed as:
\begin{equation}
    \bm{r}_i = (\bm{R} \bm{p}_i + \bm{t} - \bm{q}_i) \cdot \bm{n}_i,
\end{equation}
where \( \bm{p}_i \) and \( \bm{q}_i \) represent the matched point pairs from the current LiDAR sweep and the prior map, and \( \bm{n}_i \) denotes the associated plane normal. The optimization objective is to determine the rotation \( \bm{R} \) and translation \( \bm{t} \) that minimize the sum of squared residuals:
\begin{equation}
    (\bm{R}^*, \bm{t}^*) = \underset{\bm{R}, \bm{t}}{\arg\min} \bm{\Sigma}_{i=1}^{N} \| \bm{r}_i \|^2,
\end{equation}
% \begin{equation}
% (\bm{R}^*, \bm{t}^*) = \underset{\boldsymbol{R} \in \mathrm{SO}(3), \boldsymbol{t} \in \mathbb{R}^3}{\arg \min } \bm{\Sigma}_{i=1}^{N} \| \bm{r}_i \|^2,
% \end{equation}

Utilizing the Hessian matrix \( \bm{H} \), pose updates are computed:
\begin{equation}
    \bm{H} \begin{bmatrix}
        \Delta \bm{\theta}^\top &
        \Delta \bm{t}^\top
    \end{bmatrix}^\top
    = -\bm{J}_{\text{dm}}^\top \bm{r}_{\text{dm}},
\end{equation}
% \begin{equation}
% \begin{bmatrix}
% \Delta \bm{\theta}^\top & \Delta \bm{t}^\top
% \end{bmatrix} \bm{H}^\top = -\bm{r}_{dm}^\top \bm{J}_{dm},
% \end{equation}
% \begin{equation}
% \begin{bmatrix}
% \Delta \bm{\theta} \\
% \Delta \bm{t}
% \end{bmatrix}^\top
% \bm{H}^\top = -\bm{r}_{dm}^\top \bm{J}_{dm},
% \end{equation}
where \( \bm{J}_{\text{dm}} \) denotes the Jacobian of residuals and \( \bm{r}_{dm} \) is the residuals vector. The solution to this system provides incremental updates \( \Delta \bm{\theta} \) and \( \Delta \bm{t} \). We then compute the covariance matrix \( \bm{\Sigma}_{\text{dm}} \) as:
\begin{equation}
    \bm{\Sigma}_{\text{dm}} \approx \bm{H}^{-1}, \quad \bm{H} = \bm{J}_{\text{dm}}^\top \bm{W}_{\text{dm}} \bm{J}_{\text{dm}}.
\end{equation}
This approximation assumes Gaussian noise and a local quadratic cost function through the inverse Hessian matrix $\bm{H}$. Here, $\bm{W}_{\text{dm}}$ is the weighting matrix related to LiDAR measurement noise.

% which is computed as:
% \begin{equation}
% \bm{H} = \bm{J}_{dm}^\top \bm{W} \bm{J}_{dm},
% \end{equation}
% where \( \bm{W} \) is the weighting matrix, inversely proportional to the LiDAR measurement noise.

\subsection{Degeneracy Detection}\label{sec:enhanced_degeneration_detection}

The ICP covariance derived from the Hessian matrix is often inadequate for comprehensively identifying degenerate conditions, which are particularly prevalent in environments with repetitive or sparse features\cite{Tuna2022NovXicp}. The objective in addressing degeneration is to reliably identify degenerate directions and then mitigate their effects through optimization strategies or sensor weight adjustments. Following the methodology proposed by \cite{tagliabue2021lion}, we employ singular value decomposition (SVD) on the Hessian matrix to analyze the condition numbers continually to ascertain the principal directions of degeneracy.
\begin{equation}
    \bm{H}_{\bm{X}} = \bm{U} \bm{\Sigma} \bm{V}^\top,
\end{equation}
where \( \bm{H}_{\bm{X}} \) is the Hessian sub-matrix for rotation or translation, and SVD is indeed used to discern the singular values \( \bm{\Sigma} \) with orthogonal matrices \( \bm{U} \) and \( \bm{V} \). Due to different scale of translation and rotation part, the separate condition number \( \kappa(\bm{H}_{\bm{X}}) \) is then given as $\kappa(\bm{H}_{\bm{X}}) = \sigma_{\max} / \sigma_{\min}$, with \( \sigma_{\max} \) and \( \sigma_{\min} \) being the largest and smallest singular values. A high condition number signifies potential degeneration, indicating a need for adaptive constraints.

% 为什么这么做需要补充下
Next, the directional contributions of each correspondence pair to the pose update are then determined by $\bm{C} = -\bm{J}^\top \bm{r}$, which yields a vector \( \bm{C} \) representing the influence on the six pose parameters. We compute the relative contributions in translation and rotation by calculating the ratio of correspondences that maximally contribute to each pose parameter: $\text{CR}_{\text{dim}} = N_{\text{dim}} / N_{\text{total}}$.
% \begin{equation}
% \text{CR}_{\text{dim}} = \frac{N_{\text{dim}}}{N_{\text{total}}},
% \end{equation}
The contribution ratio \(\text{CR}_{\text{dim}}\) for a specific dimension is defined by the number of correspondences, \(N_{\text{dim}}\), with the maximal contribution in that dimension, relative to the total correspondences, \(N_{\text{total}}\). A dimension with a higher \(\text{CR}_{\text{dim}}\) demonstrates a predominant influence, directing the adjustment of constraints to effectively counteract degeneracy. This mechanism ensures precise detection and management of significant degeneracy, thus preserving the system's accuracy. In line with LOAM's\cite{Zhang2016On, Zhang2014Jun} efficiency-first approach, degeneracy detection is conducted only in the first iteration, which is an effective strategy in practical scenarios.

\section{Uncertainty Estimation in Factor Graph}
Compared to many graph-based SLAM systems that only adopt a fixed diagonal noise model \cite{Shan2021Jan, Shan2018LEGOLOAML}, they are unable to analyze the uncertainty of the final pose estimation. This section will introduce the method for estimating the uncertainty of the odom factor, then analyze how each factor in the entire factor graph affects the pose uncertainty propagation.

\subsection{Odom Factor}\label{sec:odom_cov}
Given two poses $\bm{X}_1$ and $\bm{X}_2$ with their respective covariance matrices $\bm{\Sigma}_1$ and $\bm{\Sigma}_2$ from the front-end odometry, and their cross-covariance $\bm{\Sigma}_{12}$, the joint covariance matrix is:
\begin{equation}
    \bm{\Sigma}_{\text{joint}}=\left[\begin{array}{cc}
            \bm{\Sigma}_1      & \bm{\Sigma}_{12} \\
            \bm{\Sigma}_{12}^T & \bm{\Sigma}_2
        \end{array}\right].
\end{equation}

Our goal is to estimate the covariance of the relative pose $\bm{X}_{12}=\bm{X}_1^{-1} \bm{X}_2$, which can be computed by transforming the joint distribution of $\bm{X}_1$ and $\bm{X}_2$. The Schur complement\cite{barfoot2017state} of $\bm{\Sigma}_2$ in the joint covariance matrix is calculated in (\ref{eq:schur}).
\begin{equation}\label{eq:schur}
    \bm{\Sigma}_{\text{schur}}= \bm{\Sigma}_1-  \bm{\Sigma}_{12} \bm{ \Sigma}_2^{-1} \bm{\Sigma}_{12}^T.
\end{equation}

Due to the computational challenges associated with the inversion of $\bm{\Sigma}_2$ in practical applications, we adopt a Jacobian-based approach to approximate the covariance. The full covariance propagation is computed as:
\begin{equation}
    \bm{\Sigma}_{\text{full}} \approx \bm{J}_1 \bm{\Sigma}_1 \bm{J}_1^T+ \bm{J}_1
    \bm{\Sigma}_{12} \bm{J}_2^T+\bm{J}_2 \bm{\Sigma}_{12}^T \bm{J}_1^T+\bm{J}_2 \bm{\Sigma}_2 \bm{J}_2^T.
\end{equation}

Under the assumption \cite{mangelson2020characterizing, barfoot2017state, barfoot2014associating, chen2021cramer,wu2022qpep} of independence between $\bm{X}_1$ and $\bm{X}_2$, we simplify this by taking $\bm{\Sigma}_{12}=\bm{0}$. Hence, we can use a more straightforward Jacobian-based propagation:
\begin{equation}
    \bm{\Sigma}_{\text{lo}} \approx  \bm{J}_1 \bm{\Sigma}_1 \bm{J}_1^T+\bm{J}_2 \bm{\Sigma}_2 \bm{J}_2^T.
\end{equation}

The computation of pose Jacobians in our highly linear systems is complex and challenging. We directly employ the adjoint representation for approximation. Let $\text{Ad}_{{X}_{12}}$ represent the adjoint of the relative pose $\bm{X}_{12}$, then the covariance propagation can be approximated as:
\begin{equation}
    \bm{\Sigma}_{\text{lo}} \approx \text{Ad}_{{X}_{12}} \bm{\Sigma}_1 \text{Ad}_{{X}_{12}}^T + \bm{\Sigma}_2.
\end{equation}

\subsection{Uncertainty Propagation in Factor Graph}\label{sec:uncertainty_analysis}
We represent the uncertainty of different factors through their covariance matrices, which are inverted to form the corresponding information matrices. Then the global information matrix \( \bm{\Lambda} \) is synthesized from these individual matrices and is related to the global Jacobian matrix \( \bm{J} \). For clarity, we consider a minimal case with just three poses \( \bm{X} = \{\bm{X}_1, \bm{X}_2, \bm{X}_3\} \). Incorporating various factors between these poses, as shown in Fig.\ref{fig:factor_graph}, we construct the total Jacobians and residual vectors as (\ref{eq:jac_total}).

\begin{equation}
    \label{eq:jac_total}
    \scriptsize
    % \scalebox{0.8}{ % Adjust the scale factor as needed
    \bm{J}=
    \begin{bmatrix}
        \bm{J}_{\text{lo}}^{1,2}   & -\bm{J}_{\text{lo}}^{1,2} & \bm{0}_{6 \times 6}       \\
        \bm{0}_{6 \times 6}        & \bm{J}_{\text{lo}}^{2,3}  & -\bm{J}_{\text{lo}}^{2,3} \\
        \bm{J}_{\text{dm}}^{1}     & \bm{0}_{6 \times 6}       & \bm{0}_{6 \times 6}       \\
        \bm{0}_{6 \times 6}        & \bm{J}_{\text{dm}}^{2}    & \bm{0}_{6 \times 6}       \\
        \bm{0}_{6 \times 6}        & \bm{0}_{6 \times 6}       & \bm{J}_{\text{dm}}^{3}    \\
        \bm{0}_{3 \times 6}        & \bm{J}_{\text{gf}}^{2}    & \bm{0}_{3 \times 6}       \\
        \bm{0}_{3 \times 6}        & \bm{0}_{3 \times 6}       & \bm{J}_{\text{gf}}^{3}    \\
        \bm{J}_{\text{lc}}^{1,3} & \bm{0}_{6 \times 6}       & \bm{J}_{\text{lc}}^{3,1}  \\
        \bm{0}_{6 \times 6}        & \bm{J}_{\text{nm}}^{2,3}  & -\bm{J}_{\text{nm}}^{2,3} \\
    \end{bmatrix}_{48\times18},
    \quad
    \bm{r}=
    \begin{bmatrix}
        \bm{r}_{\text{lo}}^{1,2} \\
        \bm{r}_{\text{lo}}^{2,3} \\
        \bm{r}_{\text{dm}}^{1}   \\
        \bm{r}_{\text{dm}}^{2}   \\
        \bm{r}_{\text{dm}}^{3}   \\
        \bm{r}_{\text{gf}}^{2}   \\
        \bm{r}_{\text{gf}}^{3}   \\
        \bm{r}_{\text{lc}}^{1,3} \\
        \bm{r}_{\text{nm}}^{2,3} \\
    \end{bmatrix}.
    % }
\end{equation}

% 与factor graph图对应
The optimization process seeks to minimize the squared norm of the linearized residual vector:
\begin{equation}
    \delta\bm{x}^* = \arg\min_{\delta\bm{x}} \left\lVert \bm{r} + \bm{J}\delta\bm{x} \right\rVert^2.
\end{equation}

A common approach to update the state vector involves using the inverse\cite{sol2021micro}:
\begin{equation}
    \delta\bm{x}^* = -(\bm{J}^\top \bm{J})^{-1} \bm{J}^\top \bm{r},
    \quad
    \bm{X} \leftarrow \bm{X} \oplus \delta\bm{x}^*.
\end{equation}

Considering the accurate covariance matrix \( \bm{W} \), a block-diagonal matrix of individual covariance matrices, can be defined as:
\begin{equation}
    \scalebox{0.8}{ % Adjust the scale factor as needed

    $\bm{W} = \begin{bmatrix}
            \bm{\Omega}_{\text{lo}}^{1,2} & \bm{0}_{6 \times 6}           & \bm{0}_{6 \times 6} & \cdots                        & \bm{0}_{6 \times 6}           \\
            \bm{0}_{6 \times 6}           & \bm{\Omega}_{\text{lo}}^{2,3} & \bm{0}_{6 \times 6} & \cdots                        & \bm{0}_{6 \times 6}           \\
            \vdots                        & \vdots                        & \ddots              & \vdots                        & \vdots                        \\
            \bm{0}_{6 \times 6}           & \bm{0}_{6 \times 6}           & \cdots              & \bm{\Omega}_{\text{lc}}^{1,3} & \bm{0}_{6 \times 6}           \\
            \bm{0}_{6 \times 6}           & \bm{0}_{6 \times 6}           & \cdots              & \bm{0}_{6 \times 6}           & \bm{\Omega}_{\text{nm}}^{2,3} \\
        \end{bmatrix}_{48\times48},$
        }
\end{equation}
where the zero blocks signify the independent contributions of each factor. Then the global information matrix \( \bm{\Lambda} \) is formulated as (\ref{eq:global_info}).
% $\bm{\Lambda} = \bm{J}^\top \bm{W} \bm{J}$.
\begin{equation}\label{eq:global_info}
    \bm{\Lambda} = \bm{J}^\top \bm{W} \bm{J}.
\end{equation}

Applying Cholesky decomposition\cite{ila2017slam} to \( \bm{\Lambda} \) yields the covariance matrix \( \bm{\Sigma} \):
\begin{equation}
    \bm{\Lambda} = \bm{L} \bm{L}^\top,
    \quad
    \bm{\Sigma} = \bm{L}^{-\top} \bm{L}^{-1}.
\end{equation}

In the covariance matrix  \( \bm{\Sigma} \), the diagonal elements denote variances and the off-diagonal elements represent covariances, highlighting the interdependence of state variables. Incremental algorithms such as iSAM2\cite{Kaes2011isam2} efficiently update graphs by exploiting the sparsity in \( \bm{J} \) and \( \bm{\Lambda} \), showcasing the evolving nature of uncertainty propagation in factor graphs.

\section{EXPERIMENTAL RESULTS}\label{sec:experimental_results}
We conducted a comprehensive series of real-world experiments in various indoor and outdoor environments to rigorously assess the accuracy and robustness of our proposed system.
\subsection{Experimental Setup}
\subsubsection{Hardware and Software Setup}
In addition to utilizing the FusionPortable dataset\cite{Jiao2022Mar}, our experiments were conducted using three distinct sensor combinations:
% \begin{small}
\begin{enumerate}
    \item Ouster-$128$ OS1 LiDAR with measurement noise of \SI{3}{cm}, with a 200Hz time-synchronized STIM$300$ IMU.
    \item  Pandar XT$32$ LiDAR with measurement noise of \SI{1}{cm}, combined with a 100Hz time-synchronized SBG IMU (\texttt{redbird\_01} and \texttt{parkinglot\_01}).
    \item  Pandar XT$32$ LiDAR with \SI{1}{cm} measurement noise, coupled with a 700Hz $3$DM-GQ$7$ IMU (\texttt{redbird\_02}).
\end{enumerate}
% \end{small}

We utilized two types of GT trajectories for validation:
% \begin{small}
\begin{enumerate}
    \item An indoor MOCAP system providing 120Hz, 6-DoF data with \SI{1}{\cm} accuracy.
    \item An outdoor 50Hz $3$DM-GQ$7$ INS system, achieving a best-case accuracy of \SI{1.4}{cm}.
\end{enumerate}

Prior maps were obtained using Leica MS$60$, BLK$360$ or RTC$360$ scanners. Computational experiments were conducted on a desktop with an Intel i$7$ CPU, $96$ GB DDR$4$ RAM, and $1$ TB SSD. Point cloud processing was done using the Point Cloud Library (PCL) and Open3D\cite{Zhou2018open3d}, while GTSAM\cite{frank2022gtsam} addressed the optimization challenges. Voxel filter sizes were set to \SI{0.1}{m} for LiDAR frames and \SI{0.5}{m} for prior maps.

\subsubsection{Experimental Design}
Our research entailed a comprehensive set of experiments to evaluate our method, covering Trajectory Evaluation, Map Evaluation, Degeneracy Analysis, Ablation Study, and Run-Time Analysis. We conducted trajectory assessment experiments to measure the accuracy of our algorithm against real trajectories in both indoor and outdoor settings, including degenerate scenarios. These experiments are compatible with various LIO front-ends. Recognizing the challenge of acquiring GT trajectories in practice, we also conducted map evaluation experiments to estimate the overall trajectory error. This baseline included:
% \begin{small}
\begin{enumerate}
    \item LIO: FAST-LIO2 (FL2)\footnote{\url{https://github.com/hku-mars/FAST_LIO}}, LIO-SAM (LS)\footnote{\begin{footnotesize}
                  \url{https://github.com/JokerJohn/LIO_SAM_6AXIS}
              \end{footnotesize}}, LIO-Mapping (LM)\footnote{\begin{footnotesize}
                  \url{https://github.com/hyye/lio-mapping}
              \end{footnotesize}}.
    \item Map-based Localization: HDL-Localization (HDL)\footnote{\begin{footnotesize}\url{https://github.com/koide3/hdl_localization}\end{footnotesize}}, and ICP.
    \item Prior-assisted Localization: FAST-LIO-Localization (FL2L)\footnote{\begin{footnotesize}\url{https://github.com/HViktorTsoi/FAST_LIO_LOCALIZATION}\end{footnotesize}}.
\end{enumerate}
% \end{small}

In our experiments, we utilized FL2 and LS as front-end odometries, termed PFL2 and PLS, respectively. To assess the effectiveness of the core modules of our algorithm, particularly DM and ZUPT factors, an ablation study and in-depth analyses of degenerate scenarios faced by ground robots were performed. We also conducted comprehensive comparisons to evaluate the performance of these core modules.

\begin{figure}[htbp]
    \centering
    \includegraphics[width=0.5\textwidth]{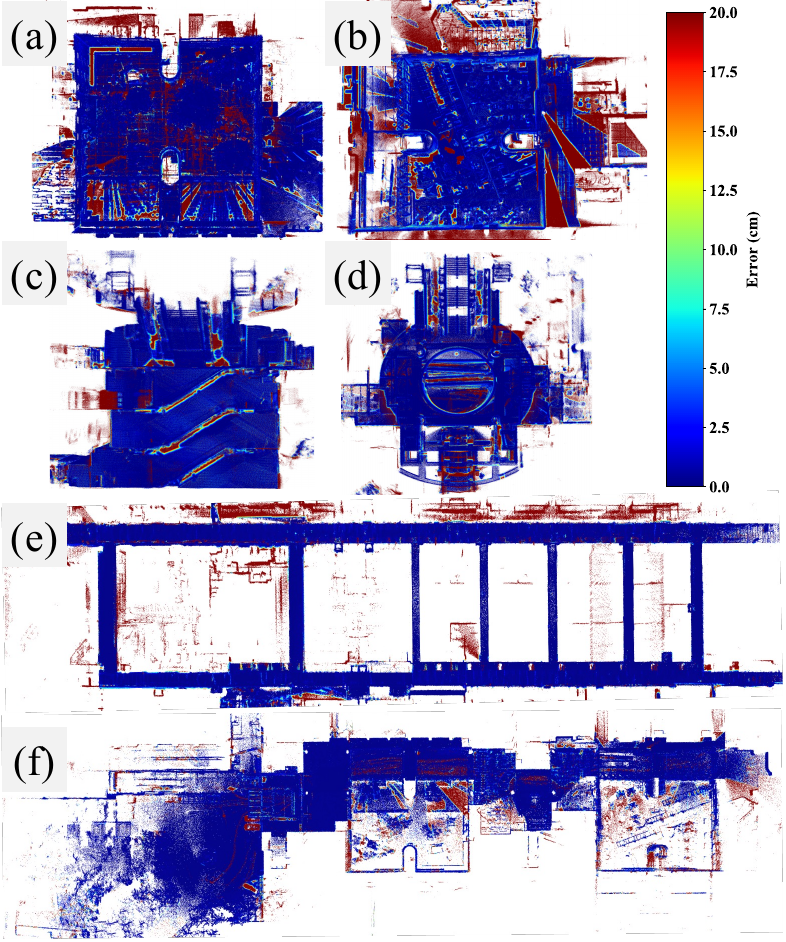}
    \caption{\begin{small} \textbf{Error Map of Diverse Campus Scenes}. The degree of color transition from blue to red indicates an increasing error in the mapped area. (a) \texttt{garden\_day} (\SI{259.5}{\m}), (b) \texttt{canteen\_day} (\SI{253.1}{\m}), (c) XZ-view of \texttt{escalator\_day} (\SI{600.1}{\m}), (d) XY-view of \texttt{escalator\_day} (\SI{600.1}{\m}) with ceiling removal, (e) \texttt{corridor\_day} (\SI{656.4}{\m}), (f) \texttt{building\_day} (\SI{717.8}{\m}). \end{small}}
    \label{fig:distance_error_map_all}
    \vspace{-2em}
\end{figure}

% 体现flexibale引用
% 硬件介绍放进来
\begin{table}[ht]
    \centering
    \caption{Quantitative Comparison of \textbf{ATE} [\si{\centi\meter}] and \textbf{RPE} [\si{\centi\meter}] Across Diverse Environments and Platforms Using Different Front-End Odometry Methods (PLS and PFL2).}
    \renewcommand\arraystretch{1.2} % Adjusts the height of table rows
    \begin{threeparttable}
        \begin{tabular}{@{}l S[table-format=3.2] S[table-format=3.2] S[table-format=3.2] S[table-format=2.1]@{}}
            \toprule
            {\textbf{Sequence}} & \multicolumn{2}{c}{\textbf{PLS}} & \multicolumn{2}{c}{\textbf{PFL2}}                                                             \\
            \cmidrule(lr){2-3} \cmidrule(lr){4-5}
                                & {\textbf{ATE} $\downarrow$}      & {\textbf{RPE} $\downarrow$}       & {\textbf{ATE} $\downarrow$} & {\textbf{RPE} $\downarrow$} \\
            \midrule
            % \multicolumn{5}{@{}l}{ \textbf{Handheld Indoor: OS-128 ($\sigma=3cm$), STIM300 IMU}} \\
            \multicolumn{5}{@{}l}{ \textbf{Handheld Indoor}}                                                                                                       \\
            \midrule[0.03cm]
            MCR\_slow           & 5.70                             & 3.89                              & 7.36                        & 2.49                        \\
            MCR\_normal         & 7.71                             & 4.95                              & 10.12                       & 3.78                        \\
            \midrule[0.03cm]
            % \multicolumn{5}{@{}l}{\textbf{Quadruped Robots Indoor: OS-128 ($\sigma=3cm$), STIM300 IMU}} \\
            \multicolumn{5}{@{}l}{\textbf{Quadruped Robots Indoor}}                                                                                                \\
            \midrule[0.03cm]
            MCR\_slow\_00       & 2.00                             & 0.90                              & 3.73                        & 0.54                        \\
            MCR\_slow\_01       & 2.93                             & 1.51                              & 2.81                        & 0.70                        \\
            MCR\_normal\_00     & 3.96                             & 2.49                              & 5.00                        & 0.72                        \\
            MCR\_normal\_01     & 3.58                             & 0.93                              & 5.23                        & 0.88                        \\
            \midrule[0.03cm]
            % \multicolumn{5}{@{}l}{\textbf{Handheld Outdoor: Pandar XT32 LiDAR ($\sigma=3cm$)}} 
            \multicolumn{5}{@{}l}{\textbf{Handheld Outdoor}}
            \\
            \midrule[0.03cm]
            % parkinglot\_01* & {--} & {--} & 28.10  & 1.46 \\
            parkinglot\_01*     & {--}                             & {--}                              & 26.10                       & 3.06                        \\
            % Parkinglot\_02* & {--} & {--} & 34.99  & 30.51 \\
            redbird\_01         & {--}                             & {--}                              & 9.98                        & 1.34                        \\
            redbird\_02         & {--}                             & {--}                              & 10.32                       & 1.34                        \\
            \bottomrule
        \end{tabular}
        \begin{tablenotes}[para,flushleft]
            {\footnotesize
                *: Degenerate scenarios. \\
                \item A dash (--) indicates sequences that were not tested.
            }
        \end{tablenotes}
    \end{threeparttable}
    \label{tab:traj_ape_rpe_half}
    \vspace{-1.5em}
\end{table}

% \begin{table}[ht]
% \centering
% \small % Reduce font size
% \caption{Quantitative Comparison of \textbf{ATE} [\si{\centi\meter}] and \textbf{RPE} [\si{\centi\meter}] Across Diverse Environments and Platforms Using Different Front-End Odometry Methods (HDL, PLS, and PFL2).}
% \renewcommand\arraystretch{1.2} % Adjusts the height of table rows
% \begin{threeparttable}
% \begin{tabular}{@{}l S[table-format=3.2] S[table-format=3.2] S[table-format=3.2] S[table-format=2.1]@{}}
% \toprule
% {\textbf{Seq.}} & \multicolumn{4}{c}{\textbf{Methods}} \\
% \cmidrule(lr){2-5}
% & {\textbf{HDL ATE}} & {\textbf{HDL RPE}} & {\textbf{PLS ATE}} & {\textbf{PLS RPE}} \\
% \midrule
% \multicolumn{5}{@{}l}{\textbf{Handheld Indoor}} \\
% \midrule[0.03cm]
% MCR\_slow & {--} & {--} & 5.70 & 3.89 \\
% MCR\_normal & {--} & {--} & 7.71 &  4.95 \\
% \midrule[0.03cm]
% \multicolumn{5}{@{}l}{\textbf{Quadruped Robots Indoor}} \\
% \midrule[0.03cm]
% MCR\_slow\_00 & {--} & {--} & 2.00 & 0.90 \\
% MCR\_slow\_01 & {--} & {--} & 2.93 & 1.51 \\
% MCR\_normal\_00 & {--} & {--} & 3.96 & 2.49 \\
% MCR\_normal\_01 & {--} & {--} & 3.58 & 0.93 \\
% \bottomrule
% \end{tabular}
% \begin{tablenotes}[para,flushleft]
% {\footnotesize
% *: Degenerate scenarios. \\
% \item A dash (--) indicates sequences that were not tested.
% }
% \end{tablenotes}
% \end{threeparttable}
% \label{tab:traj_ape_rpe_half}
% \vspace{-1.5em}
% \end{table}

\begin{table*}[ht]
    \centering
    \begin{threeparttable}
        \caption{Map Accuracy (AC) [\SI{}{\cm}] and Chamfer Distance (CD) Comparison for \SI{20}{cm} Threshold in Diverse Environments.}
        \label{tab:map_accuracy}
        \sisetup{detect-weight=true,detect-inline-weight=math}
        \begin{tabular}{l*{7}{S[table-format=2.2]S[table-format=2.2]}}
            \toprule
                           & \multicolumn{2}{c}{LM} & \multicolumn{2}{c}{LS} & \multicolumn{2}{c}{FL2} & \multicolumn{2}{c}{ICP} & \multicolumn{2}{c}{HDL} & \multicolumn{2}{c}{FL2L} & \multicolumn{2}{c}{PFL2}                                                                                                                                                 \\
            \cmidrule(lr){2-3} \cmidrule(lr){4-5} \cmidrule(lr){6-7} \cmidrule(lr){8-9} \cmidrule(lr){10-11} \cmidrule(lr){12-13} \cmidrule(lr){14-15}
            {Sequence}     & {AC $\downarrow$}      & {CD $\downarrow$}      & {AC $\downarrow$}       & {CD $\downarrow$}       & {AC $\downarrow$}       & {CD $\downarrow$}        & {AC $\downarrow$}        & {CD $\downarrow$}  & {AC $\downarrow$} & {CD $\downarrow$} & {AC $\downarrow$}  & {CD $\downarrow$}  & {AC $\downarrow$} & {CD $\downarrow$}  \\
            \midrule[0.03cm]
            \multicolumn{15}{@{}l}{\textbf{Handheld Outdoor}}                                                                                                                                                                                                                                                                                                    \\
            \midrule[0.03cm]
            garden\_day    & 4.14                   & 7.95                   & 3.94                    & 7.87                    & 5.98                    & 10.73                    & 3.64                     & 7.27               & 6.06              & 10.65             & $\underline{3.50}$ & $\underline{7.20}$ & \bfseries 3.48    & \bfseries 7.03     \\
            garden\_night  & 4.36                   & 8.08                   & 3.92                    & 7.83                    & 5.91                    & 10.78                    & $\underline{3.23}$       & $\underline{7.06}$ & 6.12              & 10.71             & 3.52               & 7.23               & \bfseries 3.19    & \bfseries 6.99     \\
            canteen\_day   & 5.65                   & 9.45                   & 5.48                    & 9.24                    & 6.32                    & 10.51                    & $\underline{4.86}$       & $\underline{8.23}$ & 6.59              & 10.63             & 5.59               & 9.02               & \bfseries 4.71    & \bfseries 8.08     \\
            canteen\_night & 5.60                   & 9.60                   & 5.29                    & 9.61                    & 6.77                    & 10.87                    & $\underline{5.07}$       & $\underline{8.42}$ & 6.93              & 10.96             & 5.56               & 8.94               & \bfseries 4.76    & \bfseries 8.16     \\

            \midrule[0.03cm]
            \multicolumn{15}{@{}l}{\textbf{Handheld Indoor}}                                                                                                                                                                                                                                                                                                     \\
            \midrule[0.03cm]
            escalator\_day & 5.40                   & 9.10                   & 8.65                    & 15.09                   & 6.92                    & 10.36                    & $\times$                 & $\times$           & 7.66              & 11.28             & $\underline{4.29}$ & $\underline{7.51}$ & \bfseries 3.88    & \bfseries 7.13     \\
            building\_day  & 10.11                  & 19.05                  & 7.65                    & 15.01                   & 6.68                    & 12.63                    & 6.72                     & 8.54               & 7.11              & 11.35             & $\underline{4.19}$ & $\underline{7.88}$ & \bfseries 4.14    & \bfseries 7.49     \\
            corridor\_day* & 7.40                   & 13.14                  & 6.24                    & 12.51                   & 7.28                    & 12.91                    & $\times$                 & $\times$           & 8.08              & 11.59             & 5.59               & 8.44               & \bfseries 3.99    & \bfseries 5.79     \\
            MCR\_slow      & 7.66                   & 13.84                  & $\underline{4.08}$      & 6.54                    & 6.19                    & 8.55                     & \bfseries 3.96           & \bfseries5.92      & $\times$          & $\times$          & $\times$           & $\times$           & 4.63              & $\underline{6.52}$ \\
            MCR\_normal    & 4.28                   & 7.92                   & $\underline{3.85}$      & $\underline{7.01}$      & $\times$                & $\times$                 & $\times$                 & $\times$           & $\times$          & $\times$          & $\times$           & $\times$           & \bfseries3.71     & \bfseries 5.86     \\
            dynamic\_02    & 9.58                   & 18.07                  & 9.78                    & 18.49                   & 9.63                    & 17.98                    & $\underline{4.13}$       & $\underline{8.09}$ & 7.44              & 10.96             & 4.62               & 8.34               & \bfseries 4.06    & \bfseries 7.82     \\
            dynamic\_04    & 9.66                   & 18.25                  & 9.81                    & 18.28                   & 9.68                    & 18.00                    & $\underline{4.30}$       & $\underline{7.90}$ & 4.35              & 11.36             & 5.06               & 8.42               & \bfseries 4.20    & \bfseries 7.67     \\
            % \midrule[0.03cm]

            % \midrule[0.03cm]
            % \multicolumn{5}{@{}l}{\textbf{Quadruped Robots Indoor}} \\
            % \midrule[0.03cm]
            % MCR\_slow\_00 & 3.23 & 8.28 & 3.66 & 7.62 & 3.53 & 7.01 & 3.22 & 6.60 & $\times$ & $\times$ & 3.72 & 7.06 & 3.59 & 6.98 \\
            % MCR\_slow\_01 & 3.41 & 8.65 & 4.70 & 9.25 & 3.52 & 6.71 & 3.29 & 6.53 & $\times$ & $\times$ & 3.72 & 6.95 & 3.62 & 6.85 \\
            % MCR\_normal\_01 & 7.25 & 12.62 & 4.64 & 8.37 & 5.48 & 8.63 & 3.82 & 6.95 & $\times$ & $\times$ & 5.99 & 7.34 & 4.05 & 7.28 \\
            % MCR\_normal\_00 & 7.25 & 12.62 & 4.04 & 7.90 & 5.37 & 8.60 & 3.82 & 6.95 & $\times$ & $\times$ & 5.15 & 7.07 & 3.49 & 6.84 \\
            \bottomrule
        \end{tabular}
        \begin{tablenotes}[para,flushleft]
            {\footnotesize
                *: Degenerate scenarios. {$\times$} signifies a failure of the algorithm on the respective datasets.\\
                \item  \textbf{Bold} indicates the best accuracy, \underline{underline} indicates the second best.
            }
        \end{tablenotes}
        \vspace{-1.5em}
        \label{tab:map_evaluation}
    \end{threeparttable}
\end{table*}

\subsubsection{Evaluation Metrics}
We employed widely used metrics such as Absolute Trajectory Error (ATE) and Relative Pose Error (RPE) for trajectory evaluation\cite{zhang2019rethinking}. For map evaluation, we utilized metrics like accuracy (AC) \cite{xu2020planar, Aanaes2016Nov}, and Chamfer distance (CD) \cite{wu2021balanced}. We set the maximum distance for corresponding point pairs to \SI{0.2}{\meter}, aligned the maps to exclude unrelated points, and used an error threshold of $\bm{\tau} = \SI{0.1}{\meter}$ for computing these evaluation metrics. Let $\mathcal{P}$ represent the point cloud sampled from the estimated map, and the GT point cloud map $\mathcal{M}$. For a point $p \in \mathcal{P}$, we define its distance to the GT map as:
\begin{equation} \label{eq:metric_p2p}
    \bm{d}(\bm{p}, \mathcal{M}) = \min_{\bm{m} \in \mathcal{M}} \|\bm{p} - \bm{m}\|.
\end{equation}

Accuracy is defined as the average Euclidean distance between estimated points and their corresponding matches in the GT map, calculated as AC = $ \frac{1}{N} \bm{\Sigma}_{i=1}^{N} [\text{dis}(\hat{p}_i, p_i) < \tau]$.
The Chamfer distance calculates the average of the squared distances between each point in one set and its closest point in the other set:
\begin{footnotesize}
    \begin{equation} \label{eq:metric_chamfer}
        CD_{\mathcal{P}, \mathcal{M}} = \frac{1}{2 \|\mathcal{P}\|} \bm{\sum}_{\bm{p} \in \mathcal{P}} \bm{d}(\bm{p}, \mathcal{M})^2 + \frac{1}{2 \|\mathcal{M}\|} \bm{\sum}_{\bm{m} \in \mathcal{M}} \bm{d}(\bm{m}, \mathcal{P})^2.
    \end{equation}
\end{footnotesize}

\subsection{Trajectory Evaluation}
% 强调flexible
Table \ref{tab:traj_ape_rpe_half} showcases the ATE and RPE of our PLS and PFL2 algorithms, tested on various platforms, including handheld and quadruped robots. This data spans both indoor and outdoor settings, underscoring the precision of our algorithm in trajectory generation. We directly obtain the trajectory ATE and RPE results of HDL for MCR-related data from \cite{Jiao2022Mar}. Compared to our proposed PFL2 and PLS, HDL frequently experiences localization failures in these quadruped robot scenarios. In the scenarios where HDL successfully runs, our PFL2 and PLS exhibit at least a 20\% improvement in both ATE and RPE accuracy compared to it. This demonstrates the robustness and high precision of our algorithm. In indoor scenarios, particularly in the MCR-related datasets confined to approximately \(\SI{8}{\m} \times \SI{5}{\m}\) rooms, the PFL2 algorithm exhibits overall trajectory accuracy below \SI{5}{\cm} for quadruped platform, despite the presence of significant near-field measurement noise from the LiDAR. On handheld platforms, the accuracy is notably lower compared to the quadruped platform. Conversely, the PLS, utilizing a different front-end odometry, shows superior precision in these feature-rich, confined spaces, highlighting the flexibility of our approach to adapt to SOTA odometry methods.  Benefiting from our factor graph framework and the loosely coupled integration of pose fusion, our system exhibits remarkable adaptability to SOTA LiDAR-based odometry methods. This versatility and practicality underline the robustness and applicability of our approach in real-world scenarios.

Outdoor experiments with PFL2 were conducted in larger, approximately \SI{1000}{\m} tracks, including both common campus pathways and areas with severe degeneration. In regular outdoor settings, PFL2 achieved an ATE of around \SI{10}{\cm} and an RPE of \SI{1.5}{\cm}. However, in degenerate scenarios, ATE accuracy significantly decreased to approximately \SI{30}{\cm}, while RPE remained relatively stable. These findings underscore the accuracy and robustness of our algorithm in diverse environments and its capacity to handle challenging scenarios, including degeneration.

\subsection{Map Evaluation} \label{sec:map_evaluation}

In practical applications without available GT trajectories, evaluating the generated trajectory presents a challenge. So we assess the accuracy of the estimated maps, using them as an indirect measure of trajectory precision. Table \ref{tab:map_evaluation} highlights the robustness and accuracy of our algorithm in various campus environments and across different platforms.

\begin{figure}
    \centering
    \includegraphics[width=0.5\textwidth]{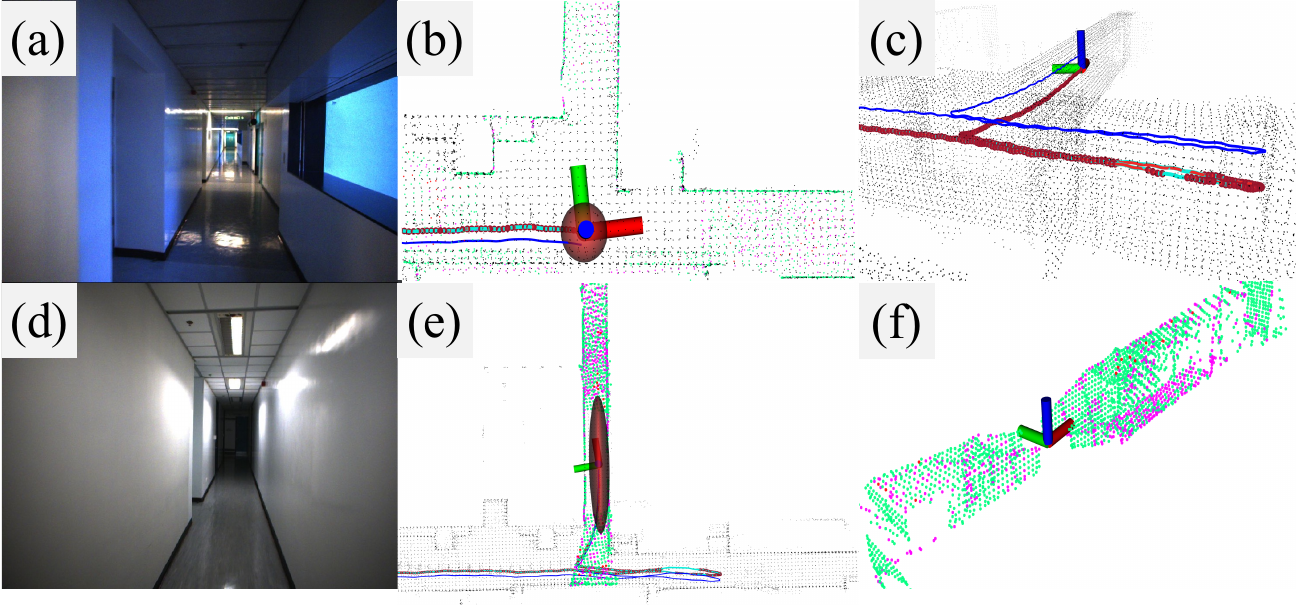}
    \caption{\begin{small} \textbf{Translation Degeneracy Analysis in Corridor}. (a) and (d) represents real-world corridor scenes. The black point cloud represents the prior map, and the red sphere with coordinate axes represents the relative constraint strength in the XYZ dimensions but is unrelated to the overall size of the ellipsoid. The flatter the ellipsoid, the more severe the degeneration in a specific dimension. The blue and light blue trajectories and the red points on the trajectories represent the FL2 odometry trajectory, our algorithm trajectory, and the pose with DM constraints. Our algorithm eliminates Z-axis drift error while ensuring robustness in a U-turn intersection (c). The point clouds of different colors in (f) indicate the corresponding number of constraints in XYZ dimensions (Section \ref{sec:enhanced_degeneration_detection}). \end{small}}
    \label{fig:degeneration_corridor}
    \vspace{-1.5em}
\end{figure}

Compared to the FL2 and other LIO methods, our PFL2 achieves a substantial improvement in map AC, with at least a 50\% enhancement. The CD is reduced by at least 30\%. Furthermore, in contrast to map-based localization algorithms, PFL2 successfully generates trajectories across all data sequences, consistently achieving near-optimal accuracy. Although it did not attain the best results in the \texttt{MCR\_normal} sequence, this is likely due to the higher noise levels in LiDAR measurements at close range. This underscores the ability of PFL2  in diverse scenarios and its superiority in leveraging map information for enhanced trajectory estimation. Our approach consistently achieves top-tier results in AC and CD metrics, outperforming SOTA techniques in demanding scenarios. Notably, the algorithm exhibits remarkable map accuracy and resilience in long-range degeneration situations, such as in the \texttt{corridor\_day} and \texttt{escalator\_day} datasets, marked by significant variations in Z-axis height (Fig.\ref{fig:degeneration_corridor}(e)).

Fig.\ref{fig:distance_error_map_all} displays the error area distribution between the estimated maps and the GT map for several sequences, applying a threshold of less than \SI{0.2}{\m}. Considering minor variations in the map and differences in scanning patterns between the LiDAR and GT Laser scanner, most areas estimated by our algorithm maintain an accuracy of nearly \SI{3}{\cm}. This level of precision, comparable to LiDAR measurement noise, indirectly verifies the accuracy and effectiveness of our trajectory estimation, supporting its use as a GT trajectory alternative.

% 联合起来说明误差和uncertainty的关系
% 为什么初始时刻精度这么差，和条件数有何关系
% 为什么退化的时候 
\begin{figure*}
    \centering
    \includegraphics[width=1.0\textwidth]{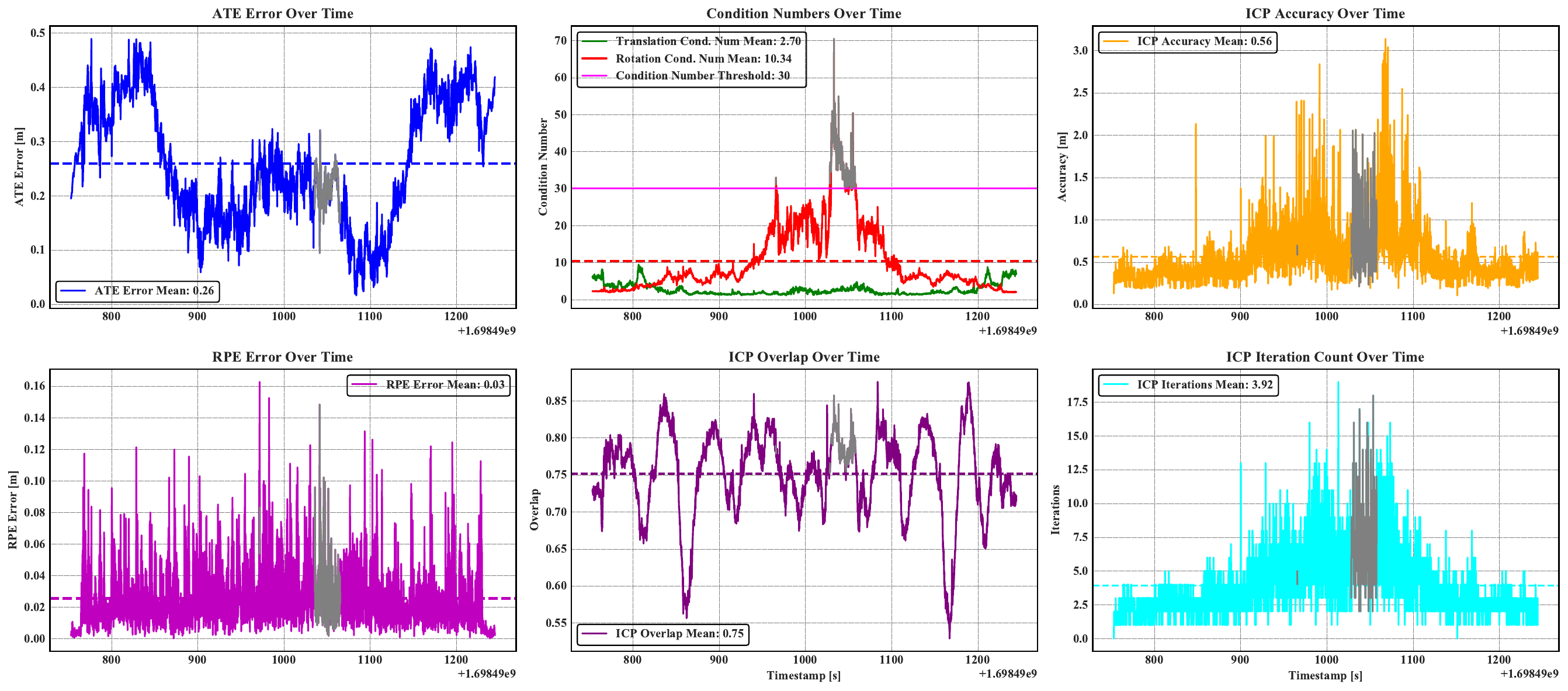}
    \caption{\begin{small} \textbf{Degeneracy Scenario Analysis}. The figure provides a detailed analysis of the DM factor in PALoc, applied to the \texttt{parkinglot\_01} dataset with yaw degeneracy. It includes a time-variant depiction of ATE, ICP accuracy, ICP iteration count, ICP overlap rate, and the condition numbers for translation and rotation. A specific threshold for the condition number, marked by the gray area, is set to identify degeneracy. It is observed that during rotational degeneracy in the scene, there is a notable decline in ICP accuracy, and ICP overlap rate, along with a significant increase in ICP iteration counts. These observations align well with the actual scenarios encountered in the parking lot setting.\end{small}}
    \label{fig:degeneracy_analysis}
    \vspace{-1.5em}
\end{figure*}

\subsection{Degeneracy Analysis}\label{sec:degeneracy_analysis}
In many ground robot scenarios, a majority of the point cloud data is constituted by ground points, effectively constraining the roll, pitch, and Z dimensions during pose estimation. This often leads to degeneracy in translation along the X\/Y-axis or rotation in the yaw direction. We thoroughly analyze this degeneracy in two distinct environments: the \texttt{corridor\_day} (X-axis degeneracy) and \texttt{parkinglot\_01} (yaw-axis degeneracy).

Fig.\ref{fig:degeneration_corridor} presents a detailed analysis of degeneracy in the \texttt{corridor\_day} scenario, especially in narrow passages (Fig.\ref{fig:degeneration_corridor}(d)), where degeneracy is identified and high-precision pose estimation is maintained. The red ellipses in the figures indicate the condition number in translation dimensions at different corridor sections. At feature-rich intersections (Fig.\ref{fig:degeneration_corridor}(a)), the constraints are well balanced as shown by the uniform ellipses in Fig.\ref{fig:degeneration_corridor}(b). However, in the narrow corridors (Fig.\ref{fig:degeneration_corridor}(e)), the elongated ellipses along the X-axis signify severe X-axis degeneracy, aligning with our theoretical analysis in Section \ref{sec:damf}. Fig.\ref{fig:degeneration_corridor}(f) investigates degeneration due to limited LiDAR points constraining the X- and Z-axes, leading to Z-axis error accumulation in narrow corridors, showcasing an improvement over the original FL2 LiDAR odometry. The blue and light blue trajectories in Fig.\ref{fig:degeneration_corridor}(c) represent the LIO and PFL2 algorithms. The effectiveness of our DM module in mitigating drift, especially in the Z-axis for LIO, is evident, corroborating the map evaluation results in Table \ref{tab:map_evaluation}.

\begin{figure}
    \centering
    \includegraphics[width=0.5\textwidth]{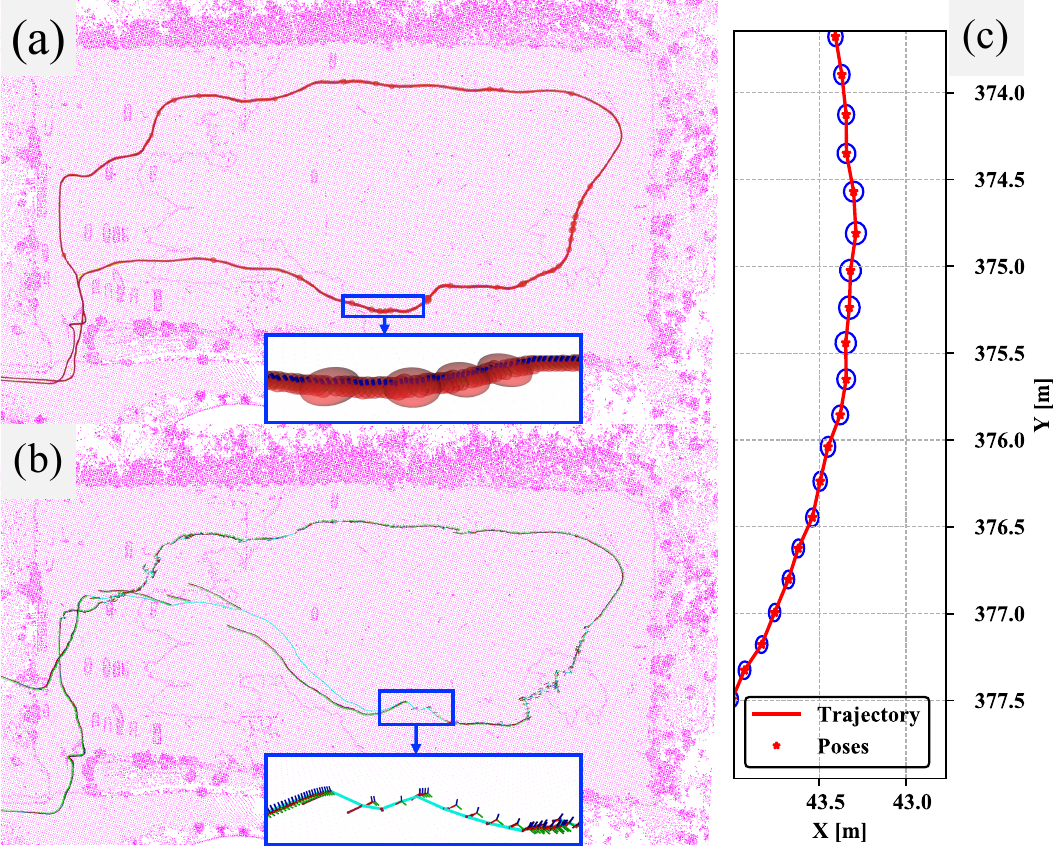}
    \caption{\begin{small} \textbf{Trajectory Comparison and Uncertainty Visualization in Rotation Degenerate Parking Lot}. (a) Demonstrates the robust performance of the PFL2 under conditions of rotational degeneracy, operating smoothly. Red spheres indicate the uncertainty in the translation part; larger spheres denote greater uncertainty, with reduced uncertainty along the Z-axis. Occasional large covariance spheres represent severe degeneracy causing matching errors, leading to the exclusion of these map factors. (b) Depicts the trajectory of FL2L in blue, with coordinate axes representing the poses, clearly showing direct localization failure in scenarios with rotational degeneracy. (c) Visualizes pose uncertainty in the X and Y dimensions at a 95\% confidence; due to the scale of the visualization, the uncertainty in Y is almost comparable to X, aligning with the observed real-world scenario. \end{small}}
    \label{fig:trajectory_with_covariances}
    \vspace{-1.5em}
\end{figure}

Fig.\ref{fig:trajectory_with_covariances} demonstrates the robustness of our PFL2 algorithm compared to FL2L using the \texttt{parkinglot\_01} dataset. Specifically, Fig.\ref{fig:trajectory_with_covariances}(c) highlights the X-Y uncertainty visualization in a targeted dataset region, where the relative size of spheres indicates their uncertainty levels. This visualization guides us in determining the reliability of different dimensions in each pose estimate, underscoring the robustness of our algorithm and how uncertainty aids in assessing pose reliability.

Fig.\ref{fig:degeneracy_analysis} presents the performance analysis of the DM module in the \texttt{parkinglot\_01} sequence using the PFL2, tracking various error metrics over time. Initially, the ATE exhibits higher values due to inaccuracies in initialization. Throughout the process, the translation condition number maintains a very low level, whereas the rotation condition number indicates severe degeneracy in the parking area, leading to similar trends in both ATE and RPE. However, degeneracy is relative, and we set a threshold of 30 for condition number detection only in the first step. This means that the periods when degeneracy is detected do not necessarily coincide with reduced point cloud matching accuracy—this would only be the case if the initial values were near perfect. Our approach is to discard the DM constraints when extreme values in the condition number are observed or when point cloud matching fails. Hence, this explains why the trends in ATE and RPE do not always align perfectly with the changes in the condition number. The data reveals how our PFL2, with its nuanced degeneracy handling, effectively manages various challenges in diverse environments.

% uncertainty 需要说明如何指导轨迹生成
% 3sigma线

% \begin{table}[htbp]
% \centering
% \footnotesize
% \caption{Ablation experiment results of \textbf{ATE} [\SI{}{\cm}], \textbf{RPE} [\SI{}{\cm}] and map \textbf{AC} [\SI{}{\cm}] on MCR\_normal\_00.}
% \renewcommand\arraystretch{1.2}
% \renewcommand\tabcolsep{15pt}
% \begin{tabular}{lccc}
% \toprule
% \textbf{Method} & \textbf{ATE} $\downarrow$ & \textbf{RPE} $\downarrow$ & \textbf{AC} $\downarrow$ \\
% \midrule[0.03cm]
% % FL2 & 9.795 & 1.163 \\
% % w/o DM & 8.391 & \textbf{1.083} \\
% % w/o LC & 5.351 & 1.580 \\
% % w/o ZV & 5.351 & 1.580 \\
% % w/o NG & 5.378 & 1.577 \\
% % All  & \textbf{5.155} & 1.508 \\
% % FL2 & 9.795 & 1.20 & 5.470 \\
% % w/o DM & 9.614 & \textbf{1.097} & 5.860 \\
% % w/o LC & 5.068 & 1.236 & \underline{4.056}\\
% % w/o NM & 5.378 & 1.577 & 4.113\\
% % % w/o ZV & 5.069 & 1.269 & 4.059\\
% % w/o GF & \underline{5.028} & 1.190 & 4.056 \\
% % All  & \textbf{5.026} & 1.189 & \textbf{4.054} \\

% FL2 & 9.80 & 1.20 & 5.47 \\
% w/o LO & 22.3 & 4.93 & 7.71\\
% w/o DM & 9.61 & 1.10 & 5.86 \\
% w/o LC & 5.47 & 0.96 & 5.38\\
% w/o NM & 5.38 & 1.58 & 4.11\\
% w/o GF & \underline{5.03} & 1.19 & 4.06 \\
% All  & \textbf{5.00} & \textbf{0.72} & \textbf{3.93} \\

% \bottomrule[0.03cm]
% \multicolumn{3}{l}{
% Note:\textbf{Bold}: best results, \underline{underlined}: second-best results.
% }\\
% \end{tabular}
% \label{tab:ablation_experiment}
% \vspace{-1em}
% \end{table}

\begin{table}[htbp]
    \centering
    \footnotesize
    \caption{Ablation study of \textbf{ATE} [\SI{}{\cm}], \textbf{RPE} [\SI{}{\cm}] and \textbf{AC} [\SI{}{\cm}] on MCR\_normal\_00.}
    \begin{threeparttable}
        \renewcommand\arraystretch{1.2}
        \renewcommand\tabcolsep{15pt}
        \begin{tabular}{lccc}
            \toprule
            \textbf{Method} & \textbf{ATE} $\downarrow$ & \textbf{RPE} $\downarrow$ & \textbf{AC} $\downarrow$ \\
            \midrule[0.03cm]
            FL2             & 9.80                      & 1.20                      & 5.47                     \\
            w/o LO          & 22.3                      & 4.93                      & 7.71                     \\
            w/o DM          & 9.61                      & 1.10                      & 5.86                     \\
            w/o LC          & 5.47                      & \underline{0.96}          & 5.38                     \\
            w/o NM          & 5.38                      & 1.58                      & 4.11                     \\
            w/o GF          & \underline{5.03}          & 1.19                      & 4.06                     \\
            All             & \textbf{5.00}             & \textbf{0.72}             & \textbf{3.93}            \\
            \bottomrule[0.03cm]
        \end{tabular}
        \begin{tablenotes}
            \small
            \item \textbf{Bold} indicates the best performance, \underline{Underlined} signifies the second-best.
        \end{tablenotes}
    \end{threeparttable}
    \label{tab:ablation_experiment}
    \vspace{-1.5em}
\end{table}

\subsection{Ablation Evaluation}
To systematically assess the contributions of different factors within our proposed system, we conducted an ablation study on the \texttt{MCR\_normal\_00}. This evaluation focused on the impact of the LO, LC, NM, GF, and DM on the overall accuracy of our system. Table \ref{tab:ablation_experiment} presents a comparative analysis of the ATE, RPE, and map AC of our proposed method when excluding these factors. Our findings reveal that the DM, GF, and NM are critical components, contributing significantly to the enhancement of trajectory accuracy. The exclusion of either factor results in a notable increase in ATE, AC, and RPE. This ablation evaluation highlights the coherent and complementary nature of the components of our proposed method, ultimately resulting in a robust and accurate system.

\begin{table}[ht]
    \centering
    \caption{Per-frame average execution time [ms] comparison of key modules on two datasets}
    \renewcommand\arraystretch{1.2}
    \begin{tabular}{l c c c c c c}
        \toprule[0.03cm]
        \textbf{Dataset} & \multicolumn{2}{c}{\textbf{DM}} & \multicolumn{2}{c}{\textbf{GO}} & \multicolumn{2}{c}{\textbf{Total}}                                                   \\
        \cline{2-7}
                         & \textbf{PFL2}                   & \textbf{ICP}                    & \textbf{PFL2}                      & \textbf{ICP}   & \textbf{PFL2}   & \textbf{ICP} \\
        \midrule[0.03cm]
        parkinglot\_01   & \textbf{205.72}                 & 289.69                          & 55.43                              & \textbf{43.13} & \textbf{261.35} & 333.01       \\
        redbird\_02      & \textbf{191.72}                 & 270.81                          & 76.46                              & \textbf{60.65} & \textbf{269.29} & 331.46       \\
        \bottomrule[0.03cm]
    \end{tabular}
    \vspace{-1.5em}
    \label{tab:algorithm_performance_comparison}
\end{table}

\subsection{Run-time Analysis}
\begin{figure}
    \centering
    \includegraphics[width=0.5\textwidth]{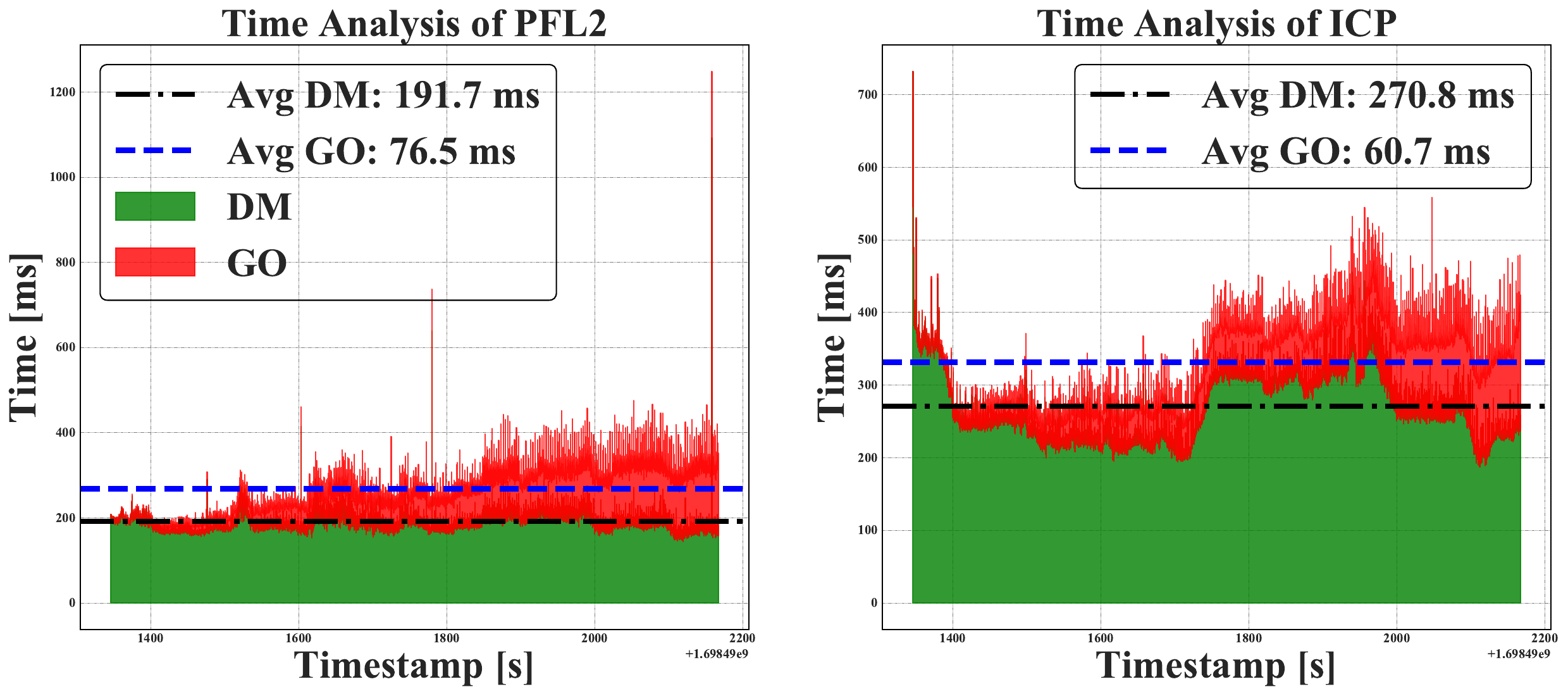}
    \caption{Performance Analysis and Comparison on \texttt{redbird\_02}.}
    \label{fig:algorithm_time_analysis}
    \vspace{-1.5em}
\end{figure}

We rigorously evaluate the performance of our proposed algorithm within extensive outdoor scenes, with a particular emphasis on the core modules that predominantly influence computational time: the DM and Graph Optimization (GO). Our DM module has been benchmarked against the Point-to-Plane ICP implementation from Open3D\cite{Zhou2018open3d}, under a strict parameter alignment such as KNN radius and maximum iterations, ensuring a fair assessment of our algorithm’s efficiency. Table \ref{tab:algorithm_performance_comparison} elucidates our findings, revealing that our DM module reduces the per-frame processing duration by approximately 30\% compared to the traditional ICP baseline, contributing to an overall minimization of at least 20\% in the average per-frame execution time. Despite this significant enhancement, the performance gains in the GO module were marginal. We attribute this to the meticulous computation of covariance for each system factor, which slightly tempered the convergence pace during optimization. Fig.\ref{fig:algorithm_time_analysis} graphically contrasts the execution times on the \texttt{redbird\_02} dataset, underscoring the enhanced performance of our algorithm over the ICP method and affirming the robustness of our approach in large-scale outdoor robotic applications.

% \begin{figure}[t]
% \setlength{\subfigcapskip}{-0.1cm}
% \setlength{\subfigbottomskip}{-0.1cm}
% \subfigure[ ]{
% \label{fig:corridor_normal}
% \includegraphics[width=.45\linewidth]{figures/algorithm_time_analysis_pfl2.pdf}
% }
% \subfigure[ ]{
% \label{fig:corridor_degenration}
% \includegraphics[width=.45\linewidth]{figures/algorithm_time_analysis_icp.pdf}
% }
% \caption{(a) Sensor configuration with corresponding coordinate frames. (b) Quadruped robot equipped with sensor suite in Motion Capture Room (MCR). (c) Prior RGB point cloud map with the estimated trajectory (red line) and map (blue point cloud) by our proposed algorithm.}
% \label{fig:sensor_kit}
% \vspace{-2em}
% \end{figure}

\subsection{Discussion}
% pose graph，所以size还行
% 图上的DM，图例线用不同标志
Our proposed trajectory generation method exhibits competitive performance across diverse scenarios and platforms. Nonetheless, several known issues warrant future research. First, the factor graph size expands as the scene enlarges, resulting in reduced system efficiency. Second, the flexibility to arbitrarily replace the front-end odometry complicates the isolated analysis of odometry degeneration. In the degeneration detection aspect, adjusting certain thresholds for various degenerate scenes is necessary. Finally, while we have estimated the uncertainty of the poses, a quantitative assessment has not been conducted, leaving the results of uncertainty lacking experimental support.

\section{Conclusion}\label{sec:conclusion}
This paper introduced a novel approach for generating dense 6-DoF GT trajectories, enhancing SLAM evaluation. We employed a factor graph optimization technique, overcoming the limitations of tracking-based methods, and effectively handling degenerate and stationary scenarios. The significant advancement of our method lies in the derivation of covariance for each factor, improving our understanding of pose uncertainty. Additionally, we integrated map evaluation criteria into this system, contributing to the precision of trajectory generation. Despite the reliance on prior maps, our open-source solution marks a notable step forward in SLAM benchmark augmentation. Future efforts will focus on expanding the applicability of our methods in diverse scenarios in robotics research.

\section*{Acknowledgments}
The authors want to express their gratitude to the editors and reviewers for their critical suggestions regarding the methodology and experiments of this study. 
They thank Bowen Yang, Xieyuanli Chen, and Binqian Jiang for their valuable corrections and insights on the grammar and logic of the paper.
Finally, they gratefully acknowledge the assistance of ChatGPT in refining the manuscript.

\bibliographystyle{IEEEtran}
\bibliography{refs.bib}

\end{document}

%% file: shortcuts.tex
\newcommand{\hide}[1]{}
\newcommand{\bdmath}{\begin{dmath}}
\newcommand{\edmath}{\end{dmath}}
\newcommand{\beq}{\begin{equation}}
\newcommand{\eeq}{\end{equation}}
\newcommand{\bdm}{\begin{displaymath}}
\newcommand{\edm}{\end{displaymath}}
\newcommand{\bea}{\begin{eqnarray}}
\newcommand{\eea}{\end{eqnarray}}
\newcommand{\beal}{\beq \begin{array}{ll}}
\newcommand{\eeal}{\end{array} \eeq}
\newcommand{\beas}{\begin{eqnarray*}}
\newcommand{\eeas}{\end{eqnarray*}}
\newcommand{\ea}{\end{array}}
\newcommand{\bit}{\begin{itemize}}
\newcommand{\eit}{\end{itemize}}
\newcommand{\ben}{\begin{enumerate}}
\newcommand{\een}{\end{enumerate}}

%% file: main.bbl
\begin{thebibliography}{10}
\providecommand{\url}[1]{#1}
\csname url@rmstyle\endcsname
\providecommand{\newblock}{\relax}
\providecommand{\bibinfo}[2]{#2}
\providecommand\BIBentrySTDinterwordspacing{\spaceskip=0pt\relax}
\providecommand\BIBentryALTinterwordstretchfactor{4}
\providecommand\BIBentryALTinterwordspacing{\spaceskip=\fontdimen2\font plus
\BIBentryALTinterwordstretchfactor\fontdimen3\font minus
  \fontdimen4\font\relax}
\providecommand\BIBforeignlanguage[2]{{%
\expandafter\ifx\csname l@#1\endcsname\relax
\typeout{** WARNING: IEEEtran.bst: No hyphenation pattern has been}%
\typeout{** loaded for the language `#1'. Using the pattern for}%
\typeout{** the default language instead.}%
\else
\language=\csname l@#1\endcsname
\fi
#2}}

\bibitem{sharp2002icp}
G.~C. Sharp \emph{et~al.}, ``Icp registration using invariant features,''
  \emph{{IEEE} Trans. Pattern Anal. Machine Intell.}, vol.~24, no.~1, pp.
  90--102, 2002.

\bibitem{Jiao2022Mar}
J.~Jiao \emph{et~al.}, ``Fusionportable: A multi-sensor campus-scene dataset
  for evaluation of localization and mapping accuracy on diverse platforms,''
  in \emph{IEEE IROS}, 2022, pp. 3851--3856.

\bibitem{Geiger2012}
A.~Geiger \emph{et~al.}, ``Are we ready for autonomous driving? the kitti
  vision benchmark suite,'' in \emph{{IEEE} Trans. Pattern Anal. Machine
  Intell.}, 2012, pp. 3354--3361.

\bibitem{Jeong2019Apr}
J.~Jeong \emph{et~al.}, ``{Complex urban dataset with multi-level sensors from
  highly diverse urban environments},'' \emph{Int. J. Robot. Res.}, vol.~38,
  no.~6, pp. 642--657, Apr. 2019.

\bibitem{Burri2016Jan}
M.~Burri \emph{et~al.}, ``{The EuRoC micro aerial vehicle datasets},''
  \emph{Int. J. Robot. Res.}, vol.~35, no.~10, pp. 1157--1163, Jan. 2016.

\bibitem{Delmerico2019May}
J.~Delmerico \emph{et~al.}, ``{Are We Ready for Autonomous Drone Racing? The
  UZH-FPV Drone Racing Dataset},'' in \emph{IEEE ICRA}.\hskip 1em plus 0.5em
  minus 0.4em\relax IEEE, May 2019, pp. 6713--6719.

\bibitem{nguyen2022ntu}
T.-M. Nguyen \emph{et~al.}, ``Ntu viral: A visual-inertial-ranging-lidar
  dataset, from an aerial vehicle viewpoint,'' \emph{Int. J. Robot. Res.},
  vol.~41, no.~3, pp. 270--280, 2022.

\bibitem{Zhang2022Aug}
L.~Zhang \emph{et~al.}, ``Hilti-oxford dataset: A millimeter-accurate benchmark
  for simultaneous localization and mapping,'' \emph{{IEEE} Robotics and
  Automation Letters}, vol.~8, no.~1, pp. 408--415, 2022.

\bibitem{ramezani2020newer}
M.~Ramezani \emph{et~al.}, ``The newer college dataset: Handheld lidar,
  inertial and vision with ground truth,'' in \emph{IEEE IROS}, 2020, pp.
  4353--4360.

\bibitem{Sier2022Oct}
H.~Sier \emph{et~al.}, ``{A Benchmark for Multi-Modal Lidar SLAM with Ground
  Truth in GNSS-Denied Environments},'' \emph{ArXiv e-prints}, Oct. 2022.

\bibitem{xu2020planar}
Q.~Xu \emph{et~al.}, ``Planar prior assisted patchmatch multi-view stereo,'' in
  \emph{AAAI}, vol.~34, no.~07, 2020, pp. 12\,516--12\,523.

\bibitem{Aanaes2016Nov}
H.~Aan{\ae}s \emph{et~al.}, ``{Large-Scale Data for Multiple-View
  Stereopsis},'' \emph{Int. J. Comput. Vision}, vol. 120, no.~2, pp. 153--168,
  Nov. 2016.

\bibitem{seb2018chall}
A.~K. Sebastian \emph{et~al.}, ``Challenges of benchmarking slam performance
  for construction specific applications,'' \emph{IEEE ICRA Workshop}, 2018.

\bibitem{Huang2010Nov}
A.~Huang \emph{et~al.}, ``{A High-Rate, Heterogeneous Data Set from the Darpa
  Urban Challenge},'' \emph{Int. J. Robot. Res.}, vol.~29, pp. 1595--1601, Nov.
  2010.

\bibitem{Peynot2010Nov}
T.~Peynot \emph{et~al.}, ``{The Marulan Data Sets: Multi-sensor Perception in a
  Natural Environment with Challenging Conditions},'' \emph{Int. J. Robot.
  Res.}, vol.~29, no.~13, pp. 1602--1607, Nov. 2010.

\bibitem{Pandey2011Mar}
G.~Pandey \emph{et~al.}, ``{Ford Campus vision and lidar data set},''
  \emph{Int. J. Robot. Res.}, vol.~30, no.~13, pp. 1543--1552, Mar. 2011.

\bibitem{Maddern2016Nov}
W.~Maddern \emph{et~al.}, ``{1 year, 1000 km: The Oxford RobotCar dataset},''
  \emph{Int. J. Robot. Res.}, vol.~36, no.~1, pp. 3--15, Nov. 2016.

\bibitem{Schubert2018Oct}
D.~Schubert \emph{et~al.}, ``{The TUM VI Benchmark for Evaluating
  Visual-Inertial Odometry},'' \emph{IEEE IROS}, pp. 1680--1687, Oct. 2018.

\bibitem{ye2020monocular}
H.~Ye \emph{et~al.}, ``Monocular direct sparse localization in a prior 3d
  surfel map,'' in \emph{IEEE Intl. Conf. on Robotics and Automation
  (ICRA)}.\hskip 1em plus 0.5em minus 0.4em\relax IEEE, 2020, pp. 8892--8898.

\bibitem{magnusson2007scan}
M.~Magnusson \emph{et~al.}, ``Scan registration for autonomous mining vehicles
  using 3d-ndt,'' \emph{J. of Field Robotics}, vol.~24, no.~10, pp. 803--827,
  2007.

\bibitem{doostmohammadian2013genericity}
M.~Doostmohammadian \emph{et~al.}, ``On the genericity properties in
  distributed estimation: Topology design and sensor placement,'' \emph{IEEE
  Journal of Selected Topics in Signal Processing}, vol.~7, no.~2, pp.
  195--204, 2013.

\bibitem{doostmohammadian2021distributed}
M.~Doostmohammadian, A.~Taghieh, \emph{et~al.}, ``Distributed estimation
  approach for tracking a mobile target via formation of uavs,'' \emph{{IEEE}
  IEEE Trans. Autom. Sci. Eng.}, vol.~19, no.~4, pp. 3765--3776, 2021.

\bibitem{Zhang2014Jun}
J.~Zhang and S.~Singh, ``Loam: Lidar odometry and mapping in real-time.'' in
  \emph{Robotics: Science and systems}, vol.~2, no.~9, 2014, pp. 1--9.

\bibitem{Zhang2015May}
J.~Zhang \emph{et~al.}, ``{Visual-lidar odometry and mapping: low-drift,
  robust, and fast},'' in \emph{IEEE ICRA}.\hskip 1em plus 0.5em minus
  0.4em\relax IEEE, May 2015, pp. 2174--2181.

\bibitem{xu2022fast}
W.~Xu \emph{et~al.}, ``Fast-lio2: Fast direct lidar-inertial odometry,''
  \emph{IEEE Trans. Robotics}, 2022.

\bibitem{ye2019tightly}
H.~Ye \emph{et~al.}, ``{Tightly Coupled 3D Lidar Inertial Odometry and
  Mapping},'' in \emph{IEEE ICRA}.\hskip 1em plus 0.5em minus 0.4em\relax IEEE,
  May 2019, pp. 3144--3150.

\bibitem{Qin2020May}
C.~Qin \emph{et~al.}, ``{LINS: A Lidar-Inertial State Estimator for Robust and
  Efficient Navigation}.''\hskip 1em plus 0.5em minus 0.4em\relax IEEE, May
  2020, pp. 8899--8906.

\bibitem{jiao2021robust}
J.~Jiao \emph{et~al.}, ``Robust odometry and mapping for multi-lidar systems
  with online extrinsic calibration,'' \emph{IEEE Trans. Robotics}, 2021.

\bibitem{huang2021bundle}
H.~Huang \emph{et~al.}, ``On bundle adjustment for multiview point cloud
  registration,'' \emph{IEEE Robotics and Automation Letters}, vol.~6, no.~4,
  pp. 8269--8276, 2021.

\bibitem{lin2022r}
J.~Lin and F.~Zhang, ``R 3 live: A robust, real-time, rgb-colored,
  lidar-inertial-visual tightly-coupled state estimation and mapping package,''
  in \emph{IEEE Intl. Conf. on Robotics and Automation (ICRA)}.\hskip 1em plus
  0.5em minus 0.4em\relax IEEE, 2022, pp. 10\,672--10\,678.

\bibitem{Rosinol20icraKimera}
A.~Rosinol \emph{et~al.}, ``Kimera: an open-source library for real-time
  metric-semantic localization and mapping,'' in \emph{IEEE ICRA}, 2020, pp.
  1689--1696.

\bibitem{Ila2010Info}
V.~Ila, J.~M. Porta, and J.~Andrade-Cetto, ``Information-based compact pose
  slam,'' \emph{IEEE Transactions on Robotics}, vol.~26, no.~1, pp. 78--93, Feb
  2010.

\bibitem{Kubelka2022Mar}
V.~Kubelka \emph{et~al.}, ``Gravity-constrained point cloud registration,'' in
  \emph{IEEE IROS}, 2022, pp. 4873--4879.

\bibitem{qin2019vins}
T.~Qin \emph{et~al.}, ``{VINS-Mono: A Robust and Versatile Monocular
  Visual-Inertial State Estimator},'' \emph{IEEE Trans. Robot.}, vol.~34,
  no.~4, pp. 1004--1020, July 2018.

\bibitem{Jiang2022Puma}
B.~Jiang and S.~Shen, ``A lidar-inertial odometry with principled uncertainty
  modeling,'' in \emph{IEEE IROS}, 2022, pp. 13\,292--13\,299.

\bibitem{tagliabue2021lion}
A.~Tagliabue \emph{et~al.}, ``Lion: Lidar-inertial observability-aware
  navigator for vision-denied environments,'' in \emph{Experimental Robotics:
  The 17th International Symposium}, 2021, pp. 380--390.

\bibitem{Tuna2022NovXicp}
T.~Tuna \emph{et~al.}, ``X-icp: Localizability-aware lidar registration for
  robust localization in extreme environments,'' \emph{{IEEE} Trans. Robotics},
  pp. 1--20, 2023.

\bibitem{jiao2021Greedy}
J.~Jiao \emph{et~al.}, ``Greedy-based feature selection for efficient lidar
  slam,'' in \emph{IEEE ICRA}, 2021, pp. 5222--5228.

\bibitem{Zhang2016On}
J.~Zhang \emph{et~al.}, ``On degeneracy of optimization-based state estimation
  problems,'' pp. 809--816, 2016.

\bibitem{Shan2021Jan}
T.~Shan \emph{et~al.}, ``{LIO-SAM: Tightly-coupled Lidar Inertial Odometry via
  Smoothing and Mapping},'' in \emph{IEEE IROS}, Jan. 2021, pp. 5135--5142.

\bibitem{Shan2018LEGOLOAML}
T.~Shan and B.~Englot, ``Lego-loam: Lightweight and ground-optimized lidar
  odometry and mapping on variable terrain,'' in \emph{IEEE IROS}, 2018, pp.
  4758--4765.

\bibitem{barfoot2017state}
T.~D. Barfoot, \emph{State estimation for robotics}, 2017.

\bibitem{mangelson2020characterizing}
J.~G. Mangelson \emph{et~al.}, ``Characterizing the uncertainty of jointly
  distributed poses in the lie algebra,'' \emph{{IEEE} Trans. Robotics},
  vol.~36, no.~5, pp. 1371--1388, 2020.

\bibitem{barfoot2014associating}
T.~D. Barfoot \emph{et~al.}, ``Associating uncertainty with three-dimensional
  poses for use in estimation problems,'' \emph{{IEEE} Trans. Robotics},
  vol.~30, no.~3, pp. 679--693, 2014.

\bibitem{chen2021cramer}
Y.~Chen \emph{et~al.}, ``Cram{\'e}r--rao bounds and optimal design metrics for
  pose-graph slam,'' \emph{IEEE Transactions on Robotics}, vol.~37, no.~2, pp.
  627--641, 2021.

\bibitem{sol2021micro}
J.~Solà \emph{et~al.}, ``A micro lie theory for state estimation in
  robotics,'' 2021.

\bibitem{ila2017slam}
V.~Ila \emph{et~al.}, ``Slam++-a highly efficient and temporally scalable
  incremental slam framework,'' \emph{Int. J. Robot. Res.}, vol.~36, no.~2, pp.
  210--230, 2017.

\bibitem{Kaes2011isam2}
M.~Kaess \emph{et~al.}, ``isam2: Incremental smoothing and mapping with fluid
  relinearization and incremental variable reordering,'' in \emph{IEEE ICRA},
  2011, pp. 3281--3288.

\bibitem{Zhou2018open3d}
Q.-Y. Zhou \emph{et~al.}, ``Open3d: A modern library for 3d data processing,''
  \emph{arXiv preprint arXiv:1801.09847}, 2018.

\bibitem{frank2022gtsam}
\BIBentryALTinterwordspacing
F.~Dellaert \emph{et~al.}, ``borglab/gtsam,'' May 2022. [Online]. Available:
  \url{https://github.com/borglab/gtsam)}
\BIBentrySTDinterwordspacing

\bibitem{zhang2019rethinking}
Z.~Zhang and D.~Scaramuzza, ``Rethinking trajectory evaluation for slam: A
  probabilistic, continuous-time approach,'' \emph{arXiv preprint
  arXiv:1906.03996}, 2019.

\bibitem{wu2021balanced}
T.~Wu \emph{et~al.}, ``Balanced chamfer distance as a comprehensive metric for
  point cloud completion,'' \emph{NIPS}, vol.~34, pp. 29\,088--29\,100, 2021.

\end{thebibliography}
